\documentclass[11pt]{article}

\usepackage[final]{acl}

\usepackage{times}
\usepackage{latexsym}

\usepackage[T1]{fontenc}
\usepackage[utf8]{inputenc}

\usepackage{microtype}
\usepackage{enumitem}
\setlist{nolistsep}

\usepackage{inconsolata}

\usepackage{graphicx}

\usepackage{comment}
\usepackage{makecell}
\usepackage{multirow}
\usepackage{adjustbox}
\usepackage{tabularx}
\usepackage{booktabs}
\usepackage{lipsum}
\usepackage{amssymb}
\usepackage[dvipsnames]{xcolor}
\title{When Trivia Is Not Trivial: Everyday Knowledge Failures in Multilingual LLMs}

\author{Anna Mosolova \quad Djamé Seddah \\
  INRIA Paris \\
  \texttt{firstname.lastname@inria.fr} \\}

\begin{document}
\maketitle
\begin{abstract}
Quiz rooms, trivia nights, and quiz shows challenge human knowledge across a wide range of topics, from canonical facts to everyday culture. In this paper, we examine whether large language models (LLMs) can perform competitively in such settings, using quiz-style questions to test them on both common and niche topics. We introduce \textsc{TriviaRoomQA}, a multilingual benchmark designed to evaluate everyday, culturally grounded, and long-tail knowledge across 288 topics. The benchmark contains 3,300 parallel multiple-choice questions in six European languages and additional 5,340 French-only questions for a more fine-grained case study. We evaluate 30 open-weight LLMs from European, Asian, and North American providers, covering models from 7 to 70B parameters. We find that models are strong on knowledge-intensive topics such as history, geography, and mathematics, but substantially weaker on everyday popular-culture topics such as celebrities, music, movies, and news. Moreover, model performance varies across languages even for the same underlying questions, suggesting that access to factual knowledge is not always language-independent. In sum, our dataset and experiments demonstrate an important knowledge gap which is not captured by existing academic-based saturated benchmarks. %OLD we propose a new dataset to study model performance across a broad mix of trivia knowledge, from common facts to cultural and niche topics, as an addition to existing academic-based benchmarks.
%All in all, our results show that trivia is not yet trivial for LLMs.
\end{abstract}

\section{Introduction}

%The release of ChatGPT in 2022 \cite{openaiIntroducingChatGPT} has brought Large Language Models (LLMs) to the general public, turning them into everyday tools for millions of users, with reports of over 2 billions messages sent per day in 2025 \cite{Chatterji2025}. While LLMs have demonstrated strong performance on standard academic benchmarks, often matching or surpassing human performance on tasks such as MMLU \citep{hendryckstest2021}, GSM8K \citep{cobbe2021trainingverifierssolvemath}, and AIME\footnote{\url{https://artificialanalysis.ai/evaluations/aime-2025}}, their capacities on culturally grounded, low-frequency, contemporary and general knowledge remain an active area of research.

Current large language models (LLMs) perform strongly on academic and reasoning-oriented benchmarks, such as MMLU \citep{hendryckstest2021}, GSM8K \citep{cobbe2021trainingverifierssolvemath}, and AIME.\footnote{\url{https://artificialanalysis.ai/evaluations/aime-2025}} However, strong performance on these benchmarks does not necessarily imply reliable knowledge of everyday facts, popular culture, contemporary events, or culturally grounded information. %In this paper, we ask whether LLMs would also be competitive contestants in quiz-room settings, where knowledge is tested across a broad range of everyday and sometimes unexpected topics.

Recent work has started to evaluate LLM knowledge beyond academic benchmarks, focusing on cultural knowledge, long-tail facts, and general factual recall.
This line of work includes a large body of papers evaluating models' knowledge of social norms and traditions in different cultures \citep{NEURIPS2024_8eb88844, fung2024massivelymulticulturalknowledgeacquisition, chiu-etal-2025-culturalbench, karmim-etal-2026-leveraging, lin2026culturallbenchmarkingmultilingualmulticultural}.\footnote{In this paper, we use the word culture to refer to \textit{the collective beliefs, values, customs, and practices that characterize a particular group or society} \citep{unescoCulture}.}
This type of knowledge, together with specialized and temporal facts, is part of low-frequency long-tail knowledge \citep{badhe2026longtailknowledgelargelanguage}, for which new benchmarks are now being constructed \citep{sun-etal-2024-head, zhu2026lpfqalongtailprofessionalforumbased}. For general knowledge evaluation, several trivia-like datasets have also been proposed for English, such as TriviaQA, Natural Questions, and SimpleQA \citep{joshi-etal-2017-triviaqa, kwiatkowski-etal-2019-natural, wei2024measuringshortformfactualitylarge}. However, these benchmarks are limited to English and do not provide fine-grained topic annotations that would allow us to analyze which groups of everyday questions are most difficult for models.

In this paper, we extend this research direction by introducing \textsc{TriviaRoomQA}, a benchmark containing typical quiz-style questions on topics covered in everyday settings. The benchmark is divided into two parts: a parallel multilingual subset and a larger French-only subset. The multilingual part of the dataset consists of 3,300 parallel questions spanning  \textbf{110 topics} and covering geographical, historical, sensory, pop-culture, and everyday-life knowledge. The questions are provided in \textbf{six languages} (English, French, Italian, Spanish, German, Dutch), enabling cross-lingual evaluation of the same underlying knowledge.

To further move beyond the English-centric paradigm, we also introduce a larger French-only evaluation set of 5,340 additional questions covering \textbf{178 additional topics}. In this work, the multilingual part allows us to compare models' knowledge across languages. The French-only part provides a deeper case study: it expands the benchmark to 8,640 French questions and covers a wider range of topics, including both general knowledge and knowledge related to French-speaking countries. We focus on French because it is widely spoken, as it is the sixth most spoken language in the world with over 334M speakers worldwide \citep{ethnologueinsightsethnologue200}, while still often appearing as a secondary language after English in LLM training data, including in multilingual or French-oriented models such as Gaperon and Lucie \citep{godey2025gaperonpepperedEnglishfrenchgenerative, gouvert2025lucie7bllmlucietraining}.

\textsc{TriviaRoomQA}, unlike prior benchmarks, stems from quiz-like questions originally designed for human knowledge assessment in quiz rooms, TV shows, and online trivia quizzes.\footnote{\url{https://www.openquizzdb.org}} Each topic in our dataset includes 30 questions about the same domain with three varying levels of difficulty. In addition to topic and difficulty annotations, we group topics into broader categories and annotate questions with the time period and geographic region to which they are most closely related. This structure allows us to both report average model performance and, more importantly, analyze how models behave across topics, categories, difficulty levels, languages, time periods, and geographic regions.

We evaluate 30 open-weight LLMs from providers in Europe, North America, and Asia, across a range of model sizes from 7 to 70B parameters. The evaluation is framed as multiple-choice question answering: models are given the question and answer candidates are scored using log-likelihood-based evaluation. 

Our analysis shows that models perform strongly on topics typically covered in widely available encyclopedic and educational sources, such as scientific and historical knowledge, while being less reliable on long-tail and everyday culture-specific topics. For example, only 4 models out of 30 correctly answer the question \textit{"What can be seen on the pin Katniss received from a merchant at the Hob?"}, while all models correctly answer  \textit{"What is the diameter of the huge golden sphere located in Auroville?"}. This shows that models can recall specific encyclopedic facts while still failing on popular-culture details that are common in quiz settings but may be less systematically represented in encyclopedic sources. While the general difficulty of long-tail knowledge has been observed in the literature, in this work, we characterize these gaps more precisely through a fine-grained topic-level analysis across languages, difficulty levels, time periods, and geographic regions. %Moreover, we observe a sharp contrast between human and model behavior: humans tend to perform better on socially grounded and pop-culture knowledge, whereas models excel in more structured factual domains such as geography and history. 

Our analysis further shows that, while humans show gradual performance degradation with increasing difficulty within a topic, LLM performance drops across all difficulty levels once a topic is outside the model's knowledge boundary. Additionally, we demonstrate that LLMs perform better on questions related to older time periods, and that some models do not answer consistently when the same question is presented in different languages.

This paper makes the following contributions:
\begin{enumerate}
    \item We introduce \textsc{TriviaRoomQA}, a human-written quiz-style benchmark for evaluating everyday, culturally grounded, and long-tail knowledge. It contains 3,300 parallel multiple-choice questions in six European languages and 5,340 additional French-only questions, annotated with 288 topics, 24 categories, 3 difficulty levels, time-period and geographic metadata. %ranging from encyclopedic knowledge to everyday culture, pop culture, food, sports, objects, news, and entertainment.
    \item We evaluate 30 open-weight LLMs from European, North American, and Asian providers, ranging from 7B to 70B parameters and show that models perform well on  scientific and encyclopedic topics, but struggle with contemporary and popular-culture questions: accuracy ranges from 0.35 to 0.64 for models with fewer than 20B parameters and from 0.41 to 0.72 for larger models. We also find that some models answer inconsistently when the same question is presented in different languages.
    \item Through a pilot human comparison study, we show that LLMs and humans behave differently on difficult topics: while human performance degrades gradually with increasing question difficulty, LLMs show similarly low performance across difficulty levels on topics where their overall accuracy is low.
\end{enumerate}

\section{Related work}

Various benchmarks have been proposed to evaluate LLM knowledge, ranging from academic and commonsense reasoning to factual and cultural knowledge.

\paragraph{World knowledge}
Most widely used benchmarks evaluate expert, academic, or science-oriented knowledge, including MMLU \citep{hendryckstest2021}, MMLU-Pro \citep{wang2024mmlu}, BIG-Bench \citep{srivastava2023imitationgamequantifyingextrapolating}, and Humanity's Last Exam \citep{Phan2026}. Other benchmarks use school or university exam questions, such as AGIEval \citep{zhong-etal-2024-agieval} or MERA \citep{fenogenova-etal-2024-mera}, or focus on commonsense understanding, e.g., HellaSwag \citep{zellers-etal-2019-hellaswag} and PIQA \citep{Bisk_Zellers_Le_bras_Gao_Choi_2020}. While these benchmarks are useful for academic abilities, they do not target everyday knowledge.

\paragraph{Trivia-style factual question answering} Several benchmarks evaluate factual knowledge in quiz-like settings. \citet{joshi-etal-2017-triviaqa} introduce TriviaQA, a reading-comprehension dataset built from trivia questions. Natural Questions dataset \citep{kwiatkowski-etal-2019-natural} evaluates models' factual knowledge using questions based on real search-engine queries, while PopQA \citep{mallen-etal-2023-trust} uses questions generated from Wikidata triples. Lastly, SimpleQA \citep{wei2024measuringshortformfactualitylarge} evaluates models' everyday knowledge on 10 topics using human-written questions.
%factual knowledge questions also exist, e.g., SimpleQA \citep{wei2024measuringshortformfactualitylarge}, PopQA \citep{mallen-etal-2023-trust} automatically generated from Wikidata, TriviaQA \citep{joshi-etal-2017-triviaqa} containing trivia questions with automatically retrieved Wikipedia articles containing the answers, ad well as Natural Questions generated from search engine queries \citep{kwiatkowski-etal-2019-natural}. 
These datasets are close to \textsc{TriviaRoomQA} as they do away with academic evaluation and concentrate on factual knowledge, however, they are primarily English-centric and are not designed for fine-grained analysis of model's knowledge by topic. In Table \ref{tab:comparison_datasets}, we compare \textsc{TriviaRoomQA} with these benchmarks.  %However, they are all exclusively on English and do not provide detailed partition of questions into topics thus not allowing to analyze on which groups of questions models fail more frequently. Additionally, LLMs already achieve 90+\% accuracy on some of these benchmarks. In contrast, TriviaRoomQA is created by humans and it provides questions in six languages with topic annotation allowing more in-depth analysis of models behavior apart from general performance on the full benchmark (see comparison in Table \ref{tab:comparison_datasets}).

\begin{table}[]
    \centering
    \begin{adjustbox}{max width = \linewidth}
    \begin{tabular}{p{2cm}p{2.2cm}cp{1.8cm}c}
    \hline
         \textbf{Benchmark} & \textbf{Questions} &  \textbf{Topics} & \textbf{Languages} & \textbf{Creation} \\
         \hline
         SimpleQA & 4,326 & 10 & English & Human \\
         PopQA & 14,267 & 16 & English & Automatic \\
         TriviaQA & 95K & - & English & Human \\
         Natural Questions (test) & 7,842 & - & English & Human \\
         Ours & 3,300 parallel, 5,340 French & 288 & 6 languages & Human \\
         \hline
    \end{tabular}
    \end{adjustbox}
    \caption{Comparison between existing trivia-style and general-knowledge benchmarks and our \textsc{TriviaRoomQA}.}
    \label{tab:comparison_datasets}
\end{table}

%the closest to ours are triviaQA, natural questions, POPQA and EntityQuestions, but they are either generated form wikipedia or already saturated. plus contain america dominated knowledge and do not offer topic distribution or very limited one

%our benchmark offers 288 different topics with questions created by humans and these topics cover multiculturual everyday intelligence 

\paragraph{Long-tail and cultural knowledge}
Several benchmarks evaluate models on long-tail, low-frequency, or domain-specific knowledge. For example, Head-to-Tail \citep{sun-etal-2024-head} studies model knowledge across most and least frequent entities from knowledge graphs, while LPFQA \citep{zhu2026lpfqalongtailprofessionalforumbased} tests models on long-tail professional knowledge from forums. \textsc{TriviaRoomQA} is related to this direction, as it includes questions about entities and events that may be rare in pre-training data, such as local news or region-specific references.

Cultural knowledge can also be viewed as a domain-specific evaluation setting \citep{badhe2026longtailknowledgelargelanguage}. Culture-oriented benchmarks primarily evaluate models' understanding of social practices, customs, values, and norms associated with different cultures. Such benchmarks include CulturalBench \citep{chiu-etal-2025-culturalbench}, BLEND \citep{NEURIPS2024_8eb88844}, CultureAtlas \citep{fung2024massivelymulticulturalknowledgeacquisition}, and LatamQA \citep{karmim-etal-2026-leveraging}. At the same time, several recent surveys on cultural evaluation \citep{adilazuarda-etal-2024-towards, zhou-etal-2025-culture, kabir-etal-2025-break, oh-etal-2025-culture, alkhamissi-etal-2026-hire} highlight that such approaches may be limited, as culture is often reduced to static facts about traditions, social preferences, or biases, commonly presented in multiple-choice format. In \textsc{TriviaRoomQA}, instead of evaluating models' understanding of cultural norms or alignment with a particular worldview, we test whether models possess everyday factual knowledge about movies, songs, events, and other topics that people may encounter through media, education, entertainment, and public life, and that together form part of everyday cultural knowledge.

\section{\textsc{TriviaRoomQA}}

\textsc{TriviaRoomQA} is a new dataset derived from quizzes freely available on the OpenQuizzDB platform\footnote{\url{https://www.openquizzdb.org}, available under the CC BY-SA 4.0 license. \textsc{TriviaRoomQA} is released under the same license and is available at: \url{https://github.com/anya-bel/triviaroomqa}/.
}, originally designed for human quiz settings. The benchmark consists of 3,300 parallel questions in French, English, Italian, Spanish, German, and Dutch, as well as 5,340 additional French-only questions (in total, 8,640 questions in French). Each question has four answer options, a short explanatory note\footnote{The explanatory note is not shown to models during evaluation, but it can be used in future experiments on model enhancement for this benchmark.}, and one of the three difficulty levels: beginner, confirmed, or expert. Examples of benchmark questions across different topics and difficulty levels are provided in Appendix~\ref{app:benchmark_examples}.

We define two evaluation settings: \textsc{TriviaRoomQA-Multi}, with 3,300 parallel questions in six languages (19,800 language-specific examples), and \textsc{TriviaRoomQA-French}, containing 8,640 French-only questions. 

The multilingual part covers 110 topics, while the full French collection covers 288 topics. We group the questions into 24 broad categories, such as history, geography, food, entertainment, music, news, sport, people, objects, tech, etc. Statistics on the number of questions and topics in each category are given in Table \ref{tab:stat_per_category}, and the full list of categories and corresponding topics is provided in Table \ref{tab:categories_topics} in Appendix \ref{app:dataset_statistics}.
General dataset statistics are provided in Table~\ref{tab:general_stats_triviaroomqa}.

\begin{table}[]
    \centering
    {\footnotesize
    \begin{tabular}{c|c}
    \hline
        \multicolumn{2}{c}{\textbf{\textsc{TriviaRoomQA-French}}} \\
        \hline
          \# of questions & 8,640 \\
          \# of topics & 288 \\
          \# of categories & 24 \\
         \hline
         \multicolumn{2}{c}{\textbf{\textsc{TriviaRoomQA-Multi}}} \\
         \hline
          \# of questions & 3,300 \\
          \# of topics & 110 \\
          \# of categories & 21 \\
         \hline
         \multirow{2}{*}{Annotation} & Topics, Categories, Level, \\
         & Continent$^\star$, Time period$^\star$ \\
         \hline
    \end{tabular}}
    \caption{Statistics and annotation information for the \textsc{TriviaRoomQA} dataset. Annotations marked with $^\star$ were produced automatically using Llama-70B-Instruct. }
    \label{tab:general_stats_triviaroomqa}
\end{table}

To estimate topic rarity, we used Infini-Gram \citep{Liu2024InfiniGram}\footnote{\url{https://infini-gram.io}} to check the frequency of topic key words in the OLMo-3-7B-Instruct training data.\footnote{We use this model as its training data is directly accessible through Infini-Gram.} We found that 21\% of the topics occur fewer than 100k times in its 6T-token corpus, suggesting that at least some topics in the benchmark are relatively infrequent in this particular corpus. We also checked whether OpenQuizzDB appears in FineWeb2 \citep{penedo2025fineweb2pipelinescale}: the full 20TB corpus contains only 214 mentions of the website, suggesting that OpenQuizzDB is not substantially represented in FineWeb2 and may be limited in CommonCrawl-derived training corpora. Another indication that the benchmark questions are not sufficiently represented in model training data is the relatively moderate performance observed across models in our experiments (Section \ref{sec:results}), although this does not provide conclusive evidence for all questions and models.

Finally, we use Llama-70B-Instruct to assign approximate geographic and temporal metadata to each question, when applicable.\footnote{The exact prompts for the continent and time-period annotation tasks are provided in Appendix~\ref{app:annotation_prompts}.} Specifically, we annotate questions with the continent and time period to which they are most closely related. These labels are intended for high-level analysis rather than as gold-standard annotations: they allow us to study broad geographic and temporal trends in model performance, but individual labels may contain errors. Statistics for these annotations are provided in Tables~\ref{tab:stat_per_continent} and~\ref{tab:stat_per_timeperiod} in Appendix~\ref{app:dataset_statistics}. Human validation of a sample of the annotations is provided in Appendix~\ref{app:human_validation}.

\section{Experimental setup}

We evaluate \textsc{TriviaRoomQA} using the Language Model Evaluation Harness library (lm-eval) \citep{eval-harness}\footnote{Experimental setup details are provided in Appendix \ref{app:exp_details}.}. We implement the benchmark as two tasks: (i) \textsc{TriviaRoomQA-Multi}, corresponding to the 3,300 parallel questions in six languages (19,800 language-specific examples), and (ii) \textsc{TriviaRoomQA-French}, corresponding to the full French collection of 8,640 questions. Both tasks are evaluated as multiple-choice question answering. For each example, the question is used as the prompt, and the four answer options are scored using conditional log-likelihood, following standard lm-eval protocol. The option with the highest score is selected as the model prediction, and we report accuracy. This evaluation avoids free-form answer generation and answer-normalization issues, and allows direct comparison across models and languages. Since each question has four answer options, the random baseline is 0.25.

All models are evaluated without a chat template in the main experiments, since chat templates were not available for all models. Results with chat templates are provided in Appendix~\ref{app:chat_template_results}.

\subsection{Models}

We evaluate 30 open-weight LLMs spanning multiple scales, languages, geographic origins, and training dataset sizes. The models range from 7B to 70B parameters: 19 models have fewer than 20B parameters, and 11 models have 20B parameters or more\footnote{All models were accessed using the Transformers library \citep{wolf-etal-2020-transformers}.}. Below, we provide the list of models used for the evaluation:

\textbf{European models}: EuroLLM-9B-Instruct and EuroLLM-22B-Instruct, Sabiá-7B, ALIA-40B-Instruct, Salamandra-7B-Instruct, Lucie-7B-Instruct, Ministral-8B-Instruct, Gaperon-8B-SFT and Gaperon-24B-SFT, Minerva-7B-Instruct, Velvet-14B, Apertus-8B-Instruct and Apertus-70B-Instruct, TildeOpen-30B,     LLaMmlein-7B, and YandexGPT-5-Lite-8B-Instruct.

\textbf{North American models}: Granite-4.1-8B and Granite-4.1-30B, Gemma-3-12B-IT and Gemma-3-27B-IT, OLMo-3-7B-Instruct and OLMo-3.1-32B-Instruct, Llama-3.1-8B and Llama-3.1-70B,  as well as Aya Expanse-8B and Aya Expanse-32B.

\textbf{Asian models}: ALLaM-7B-Instruct, LLM-JP-4-8B-Instruct, Qwen3.5-9B, and Qwen3.5-27B.

Tables ~\ref{tab:models_training_data} and~\ref{tab:model_references_urls} in the Appendix summarize the model information, i.e., training data scale, country of origin, language coverage, citations, and links.

In Appendix \ref{app:search_tool}, we additionally provide a preliminary study of Llama-3.1-8B and Qwen3.5-9B with access to a search tool. This experiment estimates the possible gains from internet access when answering quiz questions.

\section{Results and Discussion}
\label{sec:results}

We analyze model performance in the two \textsc{TriviaRoomQA} settings. 
First, we use \textsc{TriviaRoomQA-Multi} to test whether models answer the same questions consistently across languages. 
We then use the larger \textsc{TriviaRoomQA-French} subset for fine-grained analyses by category, difficulty, time period, and geographic region. 
Together, these analyses allow us to separate three sources of variation: the language of the question, the type of knowledge being tested, and the metadata associated with each question.

Throughout the section, we also compare models below and above 20B parameters to assess whether scale mitigates the observed weaknesses.

The uncertainty estimates for the main analyses are provided in Appendix~\ref{app:uncertainty}.

%As the main interest of this benchmark is not the average model performance overall, but rather its usage as a detection tool of lacunae in the models' knowledge and capacities, we first analyze the average models performance depending on the difficulty level of the questions, then per-category performance, per-language performance, per time-period performance, and per-continent performance. 

%Note that the random baseline for this dataset is 0.25.

%First, we find that \textsc{TriviaRoomQA} contains both broadly accessible and highly challenging questions. In the French collection, 362 questions are answered correctly by none of the 30 models, 1,552 by at most 5 models, and 411 by all models. This suggests that the benchmark is neither saturated nor uniformly impossible, motivating our analysis by difficulty, category, language, time period, and geographic region.

\begin{comment}
\begin{figure}[ht]
    \centering
    \includegraphics[width=\linewidth]{figs/questions_correctly.png}
    \caption{Distribution of question difficulty in \textsc{TriviaRoomQA-French} according to the number of models answering them correctly.}
    \label{fig:placeholder}
\end{figure}
\end{comment}

\subsection{Language: cross-lingual consistency remains uneven}
\begin{figure}[h]
    \centering
    \includegraphics[width=\linewidth]{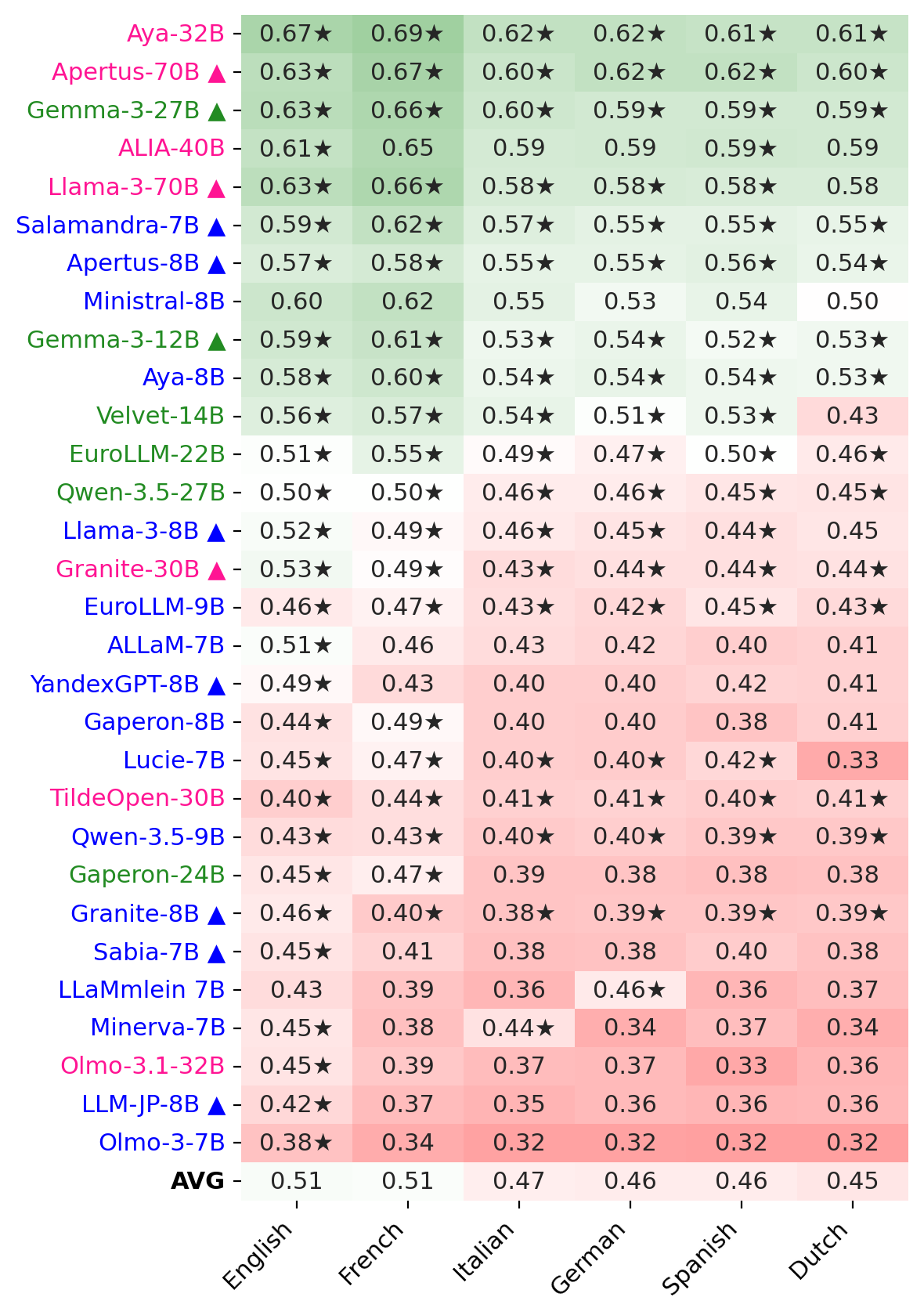}
    \caption{Model performance on the six languages of \textsc{TriviaRoomQA-Multi}. The reported metric is accuracy; greener cells indicate higher scores. %Languages are sorted from left to right by average model performance, and models are sorted from top to bottom by average performance across all languages.
    $\star$ next to a score indicates that the model was trained on that language. $\blacktriangle$ marks models trained on 10T+ tokens. 
    Model size is color-coded: \textcolor{blue}{$\leq$10B}, \textcolor{ForestGreen}{12-27B}, \textcolor{magenta}{30B+}.
    }
    \label{fig:multiling_res}
\end{figure}

Figure \ref{fig:multiling_res} presents the performance of 30 models on the six languages of \textsc{TriviaRoomQA-Multi}. 
The main observation is that models do not always answer the same questions equally well across languages. 
This is true both for models that were not trained on some of the languages in \textsc{TriviaRoomQA-Multi}, and for models whose training data includes these languages. 
This suggests that, for some models, factual knowledge is not accessed in a fully language-independent way, as also shown in prior work on multilingual factual knowledge \citep{ifergan-etal-2025-beneath}. A more detailed analysis of cross-lingual answer consistency is provided in Appendix~\ref{app:crosslingual}.

Performance varies substantially across languages and models. 
OLMo-3-7B obtains the lowest overall performance, which is expected given its English-focused training setup. 
Other models, such as Granite-8B and Llama-3.1-8B, also drop by around 0.07 accuracy points in languages other than English, despite supporting multiple languages. 
Larger models improve absolute accuracy, but do not fully solve cross-lingual inconsistency: several 20B+ models still show noticeable variation across languages.

\begin{figure*}[h]
    \centering
    \includegraphics[width=1\linewidth]{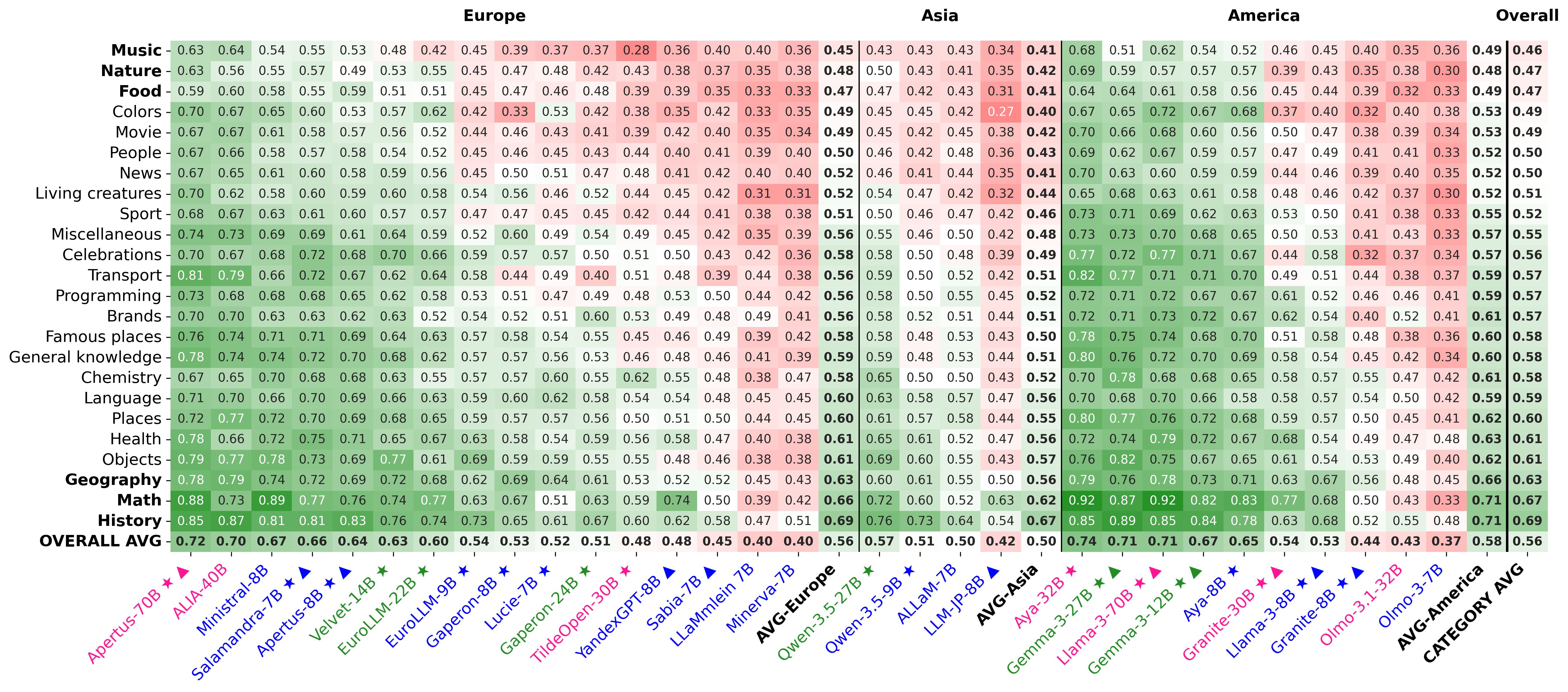}
    \caption{Model performance on \textbf{24 categories} from the \textsc{TriviaRoomQA-French}. The reported metric is accuracy; greener cells indicate higher scores. Models are sorted horizontally from left to right by performance within each region, and categories are sorted vertically from top to bottom by average performance across all models. $\star$ next to a score indicates that the model was trained on French. $\blacktriangle$ marks models trained on 10T+ tokens. 
    Model size is color-coded: \textcolor{blue}{$\leq$10B}, \textcolor{ForestGreen}{12-27B}, \textcolor{magenta}{30B+}. The category-average columns and the three highest- and lowest-performing category labels are highlighted to facilitate comparison.}
    \label{fig:category_results}
\end{figure*}

Highly multilingual models such as Salamandra-7B, Apertus-8B/70B, Gemma-3-12B/27B, and Aya-8B/32B show strong performance on all languages. 
Models trained specifically on one of the non-English benchmark languages, such as Gaperon-8B and Lucie-7B in French, LLaMmlein-7B in German, and Minerva-8B in Italian, often achieve comparable or better results in that language than in English. 
In contrast, EuroLLM and Qwen models show more uneven performance across languages in our benchmark, suggesting that cross-lingual consistency depends not only on model origin, but also on multilingual training, data scale, and model family.

%For EuroLLM, this may be related to its smaller reported training corpus compared with several other multilingual models; for Qwen, the results suggest weaker performance on this particular everyday culture-oriented evaluation.

%We can also notice that models that were specifically trained on one of the languages from the list other than English (e.g., Gaperon-8B and Lucie-7B in French, LLaMmlein-7B in German, Minerva-8B in Italian), achieve comparable or better results than in English in these languages. In general, European models show more balanced multilingual behavior, though performance still varies noticeably across languages, with English and French typically yielding the highest scores.

%More broadly, European models show relatively balanced multilingual behavior, although performance still varies noticeably across languages, with English and French typically yielding the highest scores. 
%At the same time, the pattern does not reduce to model origin alone: some non-European models, such as Aya-8B and Gemma-3-12B, also show strong multilingual robustness, while others show larger drops outside English. 

We next move to the larger French subset, where we examine whether the remaining variation is explained by topic, difficulty, time period, or geographic region.

%More broadly, European models show relatively balanced multilingual behavior, although performance still varies noticeably across languages, with English and French typically yielding the highest scores.  Smaller extra-European models more often show degradation in languages other than English, except for Aya-8B that shows strong multilingual robustness, which is consistent with its cross-lingual and multicultural training objective.

%the worst performing one is olmo which was trained on a dataset where all non-English data was filtered out. 

%european models show varying performance depending on the language with English and french being the "easiest" languages
\begin{comment}
\begin{figure*}[h]
    \centering
    \includegraphics[width=1\linewidth]{figs/category_wide_by_region_transposed_allmodels_threecolor.png}
    \caption{Model performance on \textbf{24 categories} from the \textsc{TriviaRoomQA-French}. The reported metric is accuracy; greener cells indicate higher scores. Models are sorted horizontally from left to right by performance within each region, and categories are sorted vertically from top to bottom by average performance across all models. $\star$ next to a score indicates that the model was trained on French. $\blacktriangle$ marks models trained on 10T+ tokens. 
    Model size is color-coded: \textcolor{blue}{$\leq$10B}, \textcolor{ForestGreen}{12-27B}, \textcolor{magenta}{30B+}.
    }
    \label{fig:category_results}
\end{figure*}
\end{comment}

\subsection{Category: encyclopedic knowledge is easier than popular culture}
\label{subsec:category_results}

Figure \ref{fig:category_results} presents results for 30 models on 24 categories from \textsc{TriviaRoomQA-French}. 
Performance varies substantially across categories: widely documented encyclopedic categories, such as history and geography, obtain the highest average scores, while everyday and popular-culture categories, such as music, movies, people, and news, are more difficult.

This gap is visible even for strong models. 
For example, Ministral-8B drops by 0.27 accuracy points from History to Music, and Aya-Expanse-32B drops by 0.17. 
In total, music, movies, people, and news remain below 50\% accuracy for 18 of the 30 models. 
This suggests that everyday cultural knowledge remains challenging for LLMs, possibly because such facts are less stable, less canonical, or less frequently repeated in standard factual sources than historical or geographic facts.

Models trained on more than 12T tokens, such as Salamandra, Gemma, Apertus, and Llama, achieve higher performance across many categories, and larger models such as Apertus-70B, Aya-Expanse-32B, and Granite-4.1-30B often improve on harder categories compared to their smaller counterparts. 
However, this pattern is not uniform across model families. 
In sum, scale improves absolute accuracy, but does not remove the gap between encyclopedic and everyday cultural categories.

Additional analyses in Appendices~\ref{app:category_time} and~\ref{app:detailed_error_analysis} further examine category-level differences, including their relationship with temporal variation and the topics and individual questions on which models most frequently fail.

%Models trained on more than 12T tokens, such as Salamandra, Gemma, Apertus, and Llama, tend to achieve higher performance across many categories, although training data scale might not be the only factor. 

%Among larger models, Apertus-70B, Aya-Expanse-32B, and Granite-4.1-30B improve on harder topics compared with their 8B counterparts. However, this pattern is not uniform: other model families represented at two scales show comparable performance across their smaller and larger versions. Overall, larger models often improve absolute accuracy, but the gap between encyclopedic categories and everyday cultural categories remains visible.

%all models perform well on data present in their training dataset (history, geography) from wikipedia. 

%at the same time long-tail information such as personalities remains below 50\% as well as categories that are close to human sensory experience such as music, food, and colors.

%additionaly we can see that models trained on bigger datasets (salamadra, gemma, apertus) perform much better than models with smaller training tokens number. the only exception is Qwen which shows mediocre results across almost all topics apart from the ones where all models excel independent of their training tokens number.

\subsection{Difficulty level: humans and LLMs show different patterns on unfamiliar topics}

\begin{figure}[h]
    \centering
    \includegraphics[width=0.7\linewidth]{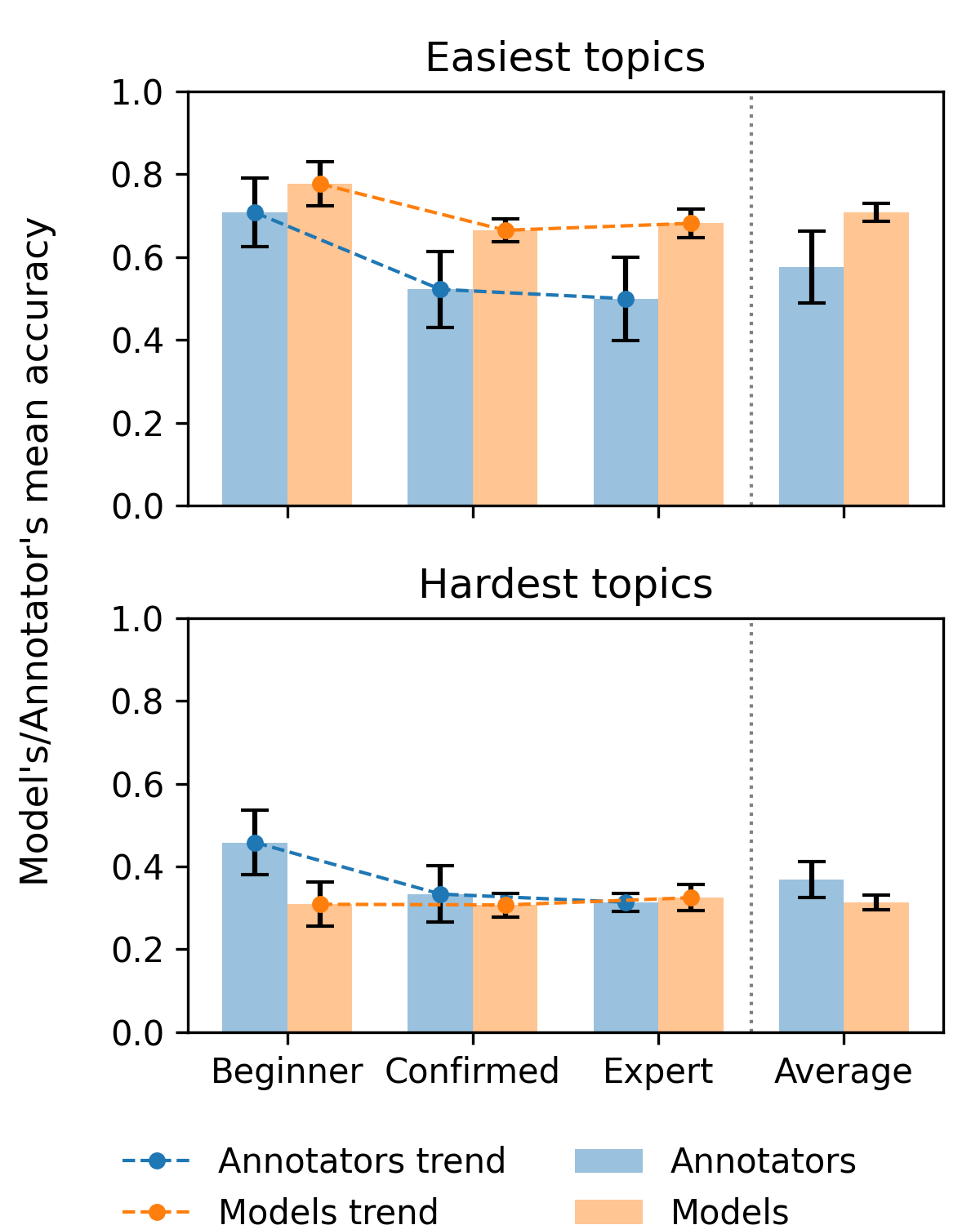}
    \caption{Average human and model performance on the six easiest and six hardest topics for models in \textsc{TriviaRoomQA-French}, grouped by question difficulty.}
    \label{fig:human_vs_llms_12topics}
\end{figure}

\begin{comment}
\begin{table}[ht]
    \centering
    \begin{adjustbox}{max width = \linewidth}
    \begin{tabular}{ccccc}
    \hline
         & \textbf{Beginner} & \textbf{Confirmed} & \textbf{Expert} & \textbf{Avg.} \\
         \hline
         Acc. & 0.57 & 0.49 & 0.45 & 0.50 \\
         \hline
    \end{tabular}
    \end{adjustbox}
    \caption{Average accuracy across 19 models on \textsc{TriviaRoomQA-French} divided according to the difficulty level of the questions as well as the overall average accuracy.}
    \label{tab:res_per_difficulty}
\end{table}
\end{comment}

On \textsc{TriviaRoomQA-French}, model performance decreases with the annotated difficulty level: average accuracy is 0.60 on beginner questions, 0.52 on confirmed questions, and 0.48 on expert questions, with an average of 0.53. 
This shows that the difficulty annotations are reflected in model performance, while also confirming that the benchmark remains challenging overall. 
Per-model results are provided in Figure~\ref{fig:detailed_niveau_perf} in the Appendix.
%In Table \ref{tab:res_per_difficulty}, we present the average results of 19 models on the French collection of the \textsc{TriviaRoomQA-French} divided by the difficulty level of the question (see Table \ref{fig:detailed_niveau_perf} in the Appendix for per-model result). 
%The results show that this benchmark is hard for the models in general, as the average performance is 50\% and also it shows that models start making more errors when the difficulty level increases. 

To further examine whether models and humans react similarly to difficulty, we selected the six easiest and six hardest topics for models, according to their average performance, and asked five French speakers to answer the corresponding questions.\footnote{The selected topics and participant information are provided in Appendix \ref{app:human_eval}.} 
Figure~\ref{fig:human_vs_llms_12topics} compares the average performance of humans and models on these two groups of topics across difficulty levels.

On the easiest topics, LLMs outperform humans on average, but the situation is reversed on the hardest topics, where humans achieve higher scores. 
More importantly, humans show a gradual decrease in performance from beginner to expert questions, while models, once they are unfamiliar with a topic, remain close to the random baseline across all difficulty levels. 
This suggests that the difficulty scale for models is not the same as for humans, with model scale potentially related to the density of topic-related information in models' training data.

\subsection{Time period: historical and timeless questions are easier for models}

\begin{figure}[!ht]
    \centering
    \includegraphics[width=1\linewidth]{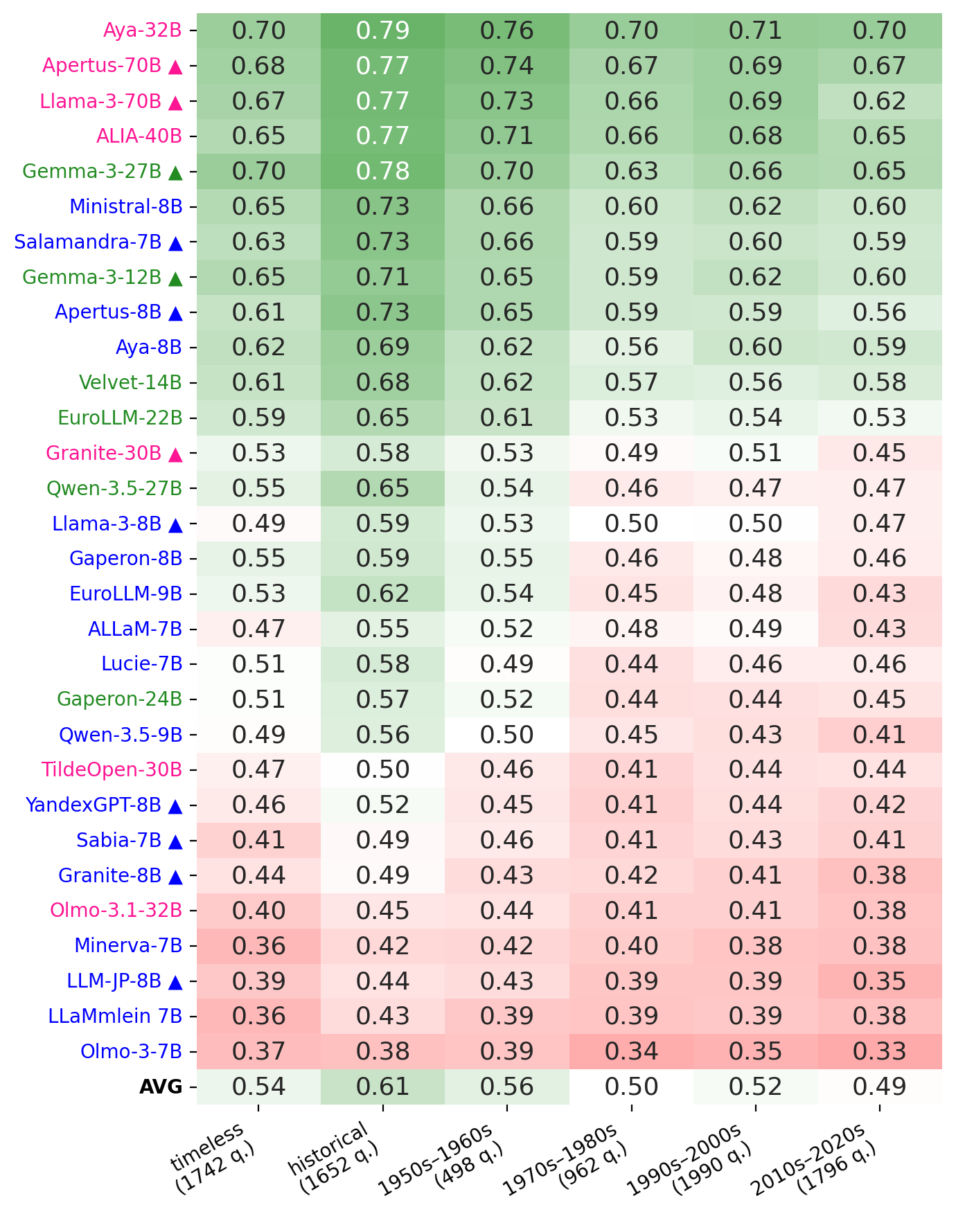}
    \caption{Model performance on \textsc{TriviaRoomQA-French} grouped by the \textbf{time period} to which each question is most closely related, when applicable. The reported metric is accuracy, greener cells indicate higher scores. $\blacktriangle$ marks models trained on 10T+ tokens. 
    Model size is color-coded: \textcolor{blue}{$\leq$10B}, \textcolor{ForestGreen}{12-27B}, \textcolor{magenta}{30B+}.
    The full results %, including more fine-grained time periods 
    are provided in Figure~\ref{fig:timeperiod_perf} in the Appendix.}
    \label{fig:timeperiod_perf_merged}
\end{figure}

Figure \ref{fig:timeperiod_perf_merged} shows model performance on \textsc{TriviaRoomQA-French} grouped by the time period most closely associated with each question, including historical questions, recent questions, and timeless questions, i.e., questions that do not depend on a specific temporal frame.

Across models, historical and timeless questions obtain the highest scores, while performance decreases for more recent periods, especially for questions from the 2000-2020s. This decrease is less pronounced for several larger models, which maintain relatively strong performance on recent questions, nevertheless the temporal gap remains visible.  One possible explanation is that older and timeless facts are more likely to appear repeatedly in stable sources such as encyclopedias, educational materials, and web pages. Recent questions, by contrast, often concern news or media events that may be less consistently represented in pre-training data. This trend is aligned with the category results, since recent questions include popular culture and news topics, which we showed in Section \ref{subsec:category_results} to be among the lowest-performing categories.

%Across all models, historical and timeless questions obtain the highest scores,  to be the easiest ones for the models, which might be related to the fact that this information is one of the main types of information present in the training data e.g., Wikipedia. After this, the closer the questions become to nowadays, the worse is the models performance probably due to the fact that new information is not still getting repeating many times. additionally, 2000-2020 period contains questions on news which are typically not part of the training dataset for the LLMs. 

\subsection{Continent: performance depends on model origin and question mix}

\begin{figure}[h]
    \centering
    \includegraphics[width=\linewidth]{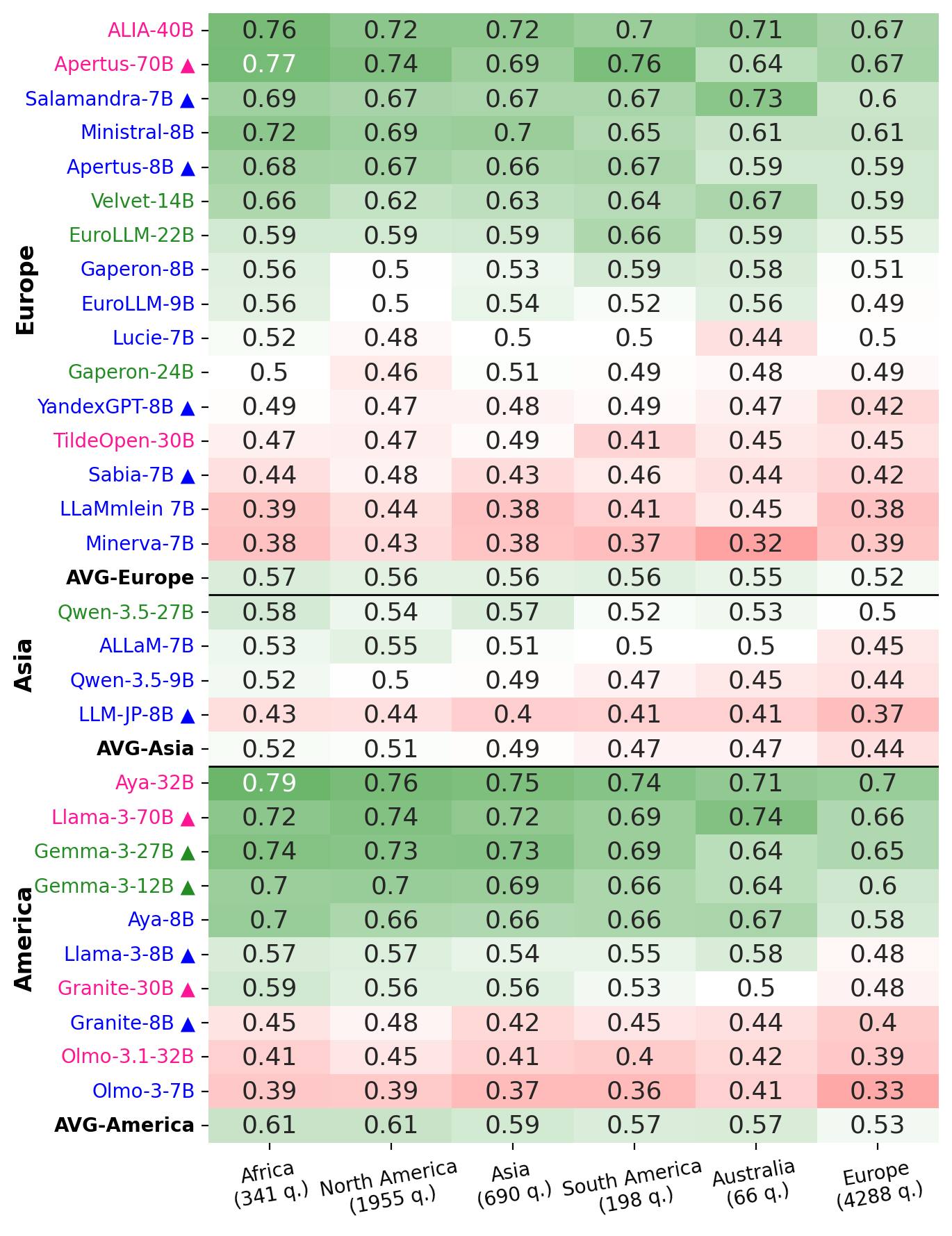}
    \caption{Model performance on \textsc{TriviaRoomQA-French}, grouped by the \textbf{continent} to which each question is most closely related, when applicable. Models are grouped by continent of origin, with the average performance for each group reported in the corresponding \textbf{AVG-\{continent\}} row. The reported metric is accuracy; greener cells indicate higher scores. $\blacktriangle$ marks models trained on 10T+ tokens. 
    Model size is color-coded: \textcolor{blue}{$\leq$10B}, \textcolor{ForestGreen}{12-27B}, \textcolor{magenta}{30B+}.
    The full results
    %, including the remaining continents and questions for which continent is not applicable, 
    are provided in Figure~\ref{fig:continent_perf_full} in the Appendix.}
    \label{fig:continent_perf}
\end{figure}

%gemma is the strongest model on all continents
Figure~\ref{fig:continent_perf} reports model performance on questions grouped by the continent to which they are most closely related. For models below 20B parameters, average performance by model origin broadly aligns with the question region: European models perform best on Europe-related questions, Asian models on Asia-related questions, and North American models on North America-related questions. 
This trend becomes less clear when larger models are included, since most 20B+ models in our evaluation come from North America and also have higher overall accuracy. 
This suggests that origin alone cannot explain the continent-level patterns.
%Therefore, continent-level differences should not be interpreted as origin effects alone, they also reflect model size, training data scale, language coverage, and question composition.

%Among individual models of fewer than 20B parameters, Gemma-3-12B achieves the highest performance across all continents, indicating strong and balanced performance across geographic groups. Ministral-8B, Salamandra-7B, Aya-Expanse-8B, Apertus-8B, and Velvet-14B also perform strongly, although with more variation depending on the continent. Gaperon-8B, EuroLLM-9B, and Llama-3.1-8B show more moderate performance across regions. Among larger models, most achieve strong performance across all continents. The main exceptions are EuroLLM-22B, Gaperon-24B, TildeOpen-30B, and OLMo-3.1-32B. One possible explanation is training data scale: these models are trained on 2-6T tokens, whereas several other 20B+ models in our evaluation are trained on larger corpora.

%For models below 20B parameters, average performance by model origin broadly aligns with the question region: European models perform best on Europe-related questions, Asian models on Asia-related questions, and North American models on North America-related questions. This trend becomes less clear when larger models are included, since most 20B+ models in our evaluation come from North America and also have higher overall accuracy. Therefore, continent-level differences should not be interpreted as origin effects alone; they also reflect model size, training data scale, language coverage, and question composition.

Among individual models, EuroLLM-22B and Qwen3.5-27B show more uneven regional performance than several similarly sized multilingual models. For EuroLLM, this may be related to its smaller reported training corpus, while for Qwen the results suggest weaker performance on this particular everyday-knowledge setting. 

Regional score differences appear to be driven by several factors.
North America-related questions obtain high scores across most models, consistent with prior work reporting stronger model performance on North American cultural knowledge \citep{chiu-etal-2025-culturalbench}. 
This tendency is visible even for LLaMmlein-7B, a German-only model trained from scratch, suggesting that the effect cannot be attributed only to English-language training data. 
Rather, it may reflect the broad coverage of U.S.-related information in web data, multilingual media, and encyclopedic sources. 
Similarly, the relatively strong scores on Africa-related questions can be partially explained by the topics covered in this subset, which include well-known encyclopedic topics such as Ancient Egypt and famous African cities. 
In contrast, Europe-related questions include more region-specific and long-tail topics, such as French cheeses, long-forgotten comedians, or Belgian beers. 
Thus, continent-level performance reflects both geographic knowledge coverage and the difficulty and topic distribution of the questions in each region.

\subsection{Effect of model size: scale helps but does not close the gaps}
The previous subsections show that larger models generally improve absolute accuracy, but do not eliminate the main performance gaps revealed by \textsc{TriviaRoomQA}. In particular, models with more than 20B parameters often rank among the strongest systems, but they still show cross-lingual inconsistency, lower performance on popular-culture and everyday categories, and sensitivity to the temporal and geographic composition of the questions.

\section{Conclusion}

In this paper, we introduce \textsc{TriviaRoomQA}, a multilingual benchmark for evaluating LLMs in settings inspired by quiz games. 
%It contains 3,300 parallel questions in six languages and 5,340 additional French-only questions targeting everyday, culturally grounded, and long-tail knowledge across a wide range of topics.

%In this paper, we introduce \textsc{TriviaRoomQA}, a multilingual benchmark for evaluating LLMs in settings inspired by human quiz rooms and trivia games. Unlike academic and science-oriented benchmarks, \textsc{TriviaRoomQA} targets everyday, culturally grounded, and long-tail knowledge across a wide range of topics. The benchmark contains 3,300 parallel questions in six languages and 5,340 additional French-only questions for more fine-grained analysis.%. We show that, despite the fact that contemporary models achieve near-perfect performance on existing science-oriented benchmarks (e.g., MMLU and MMLU-Pro \citealt{hendryckstest2021, wang2024mmlu}), they still fail to correctly answer questions on popular culture. 

Our results show that trivia remains challenging for open-weight LLMs. Models perform well on encyclopedic categories such as history and geography, but struggle with popular culture and everyday categories, such as music, movies, and news. They also answer the same questions inconsistently across languages, suggesting that factual knowledge is not always accessed in a language-independent way, as also shown in prior work \citep{ifergan-etal-2025-beneath}. 
Larger models generally improve absolute accuracy, but do not eliminate these gaps: models above 20B parameters still show cross-lingual inconsistency and weaker performance on everyday cultural knowledge.

Our benchmark also reveals that what models find difficult does not always align with what humans find difficult.  In our pilot human comparison, human performance decreases gradually as question difficulty increases, while model performance remains close to the random baseline across difficulty levels on topics where models perform poorly overall. We also find that historical and timeless questions are easier on average than recent questions, and that North America-related questions obtain high scores across most models, although continent-level results partly reflect the topic composition of each regional subset.

Finally, a preliminary experiment with search-augmented models suggests that some errors come from information that is missing from, or difficult to retrieve from, model parameters alone, since tested models improve when given access to external search.

In summary, \textsc{TriviaRoomQA} provides a fine-grained benchmark for identifying gaps in the factual knowledge stored by LLMs and in the way this knowledge is accessed across languages, topics, time periods, and geographic regions. Future work can use the benchmark to evaluate retrieval-augmented or tool-using systems, where models have access to external information rather than relying only on parametric knowledge.

%We invite everyone to use this new dataset as one more gate towards understanding what kind of knowledge models store and how they access it. 

%The next step in using this benchmark is testing it in conditions where LLMs have access to the information obtained from Retrieval-Augmented Generation or using tools such as web search. (cite papers that already use quizzes to eval rag)

\begin{comment}

problem with agentic: no internet on the server, how to host model for eval and to make queries?

todo:
\begin{itemize}
    %\item eval of 24 72 apertus and other + something multilignual
    \item divide dataset to use part of it as instruct (25\%?)
    \item SFT on yahoo
    \item apply chat template
\end{itemize}
\end{comment}

\section*{Limitations}

Despite the broad coverage of \textsc{TriviaRoomQA}, this work has several limitations. First, the multilingual part of the benchmark currently covers only six European languages, and the fine-grained analysis relies mainly on the French subset. Second, we evaluate only open-weight models, so the results may not reflect the behavior of closed commercial LLMs. Third, the continent and time-period annotations are produced automatically and should be interpreted as approximate labels rather than gold annotations. %Fourth, the search-tool experiment in Appendix~\ref{app:search_tool} is preliminary and includes only two models. 
Fourth, although OpenQuizzDB states that its questions are written by contributors following the platform's guidelines and reviewed by project members before publication, we do not independently verify the factual validity of every question.
Finally, our human comparison study involves only five participants and should be interpreted as a pilot analysis rather than a definitive estimate of human quiz performance.

\section{Ethical considerations}
\textsc{TriviaRoomQA} is derived from OpenQuizzDB, which is distributed under a CC BY-SA 4.0 license. We follow the license terms and preserve attribution to the original source. The benchmark contains trivia-style questions about public factual knowledge and does not require collecting personal information. Some topics in the dataset include questions related to alcohol and sexuality. This benchmark is intended for diagnostic research on LLM factual knowledge, and its scores should not be interpreted as a general measure of model intelligence or cultural competence.

The human comparison study involved five French-speaking participants. Participants were informed about the purpose of the study, answered voluntarily, and only average accuracy statistics are reported. We do not release individual-level responses. 

Metadata annotations (continent and time period information) are produced automatically with Llama-70B-Instruct, therefore, they may contain errors and should be used only for trend-level analysis rather than as gold-standard annotations.

\section*{Acknowledgments}

We warmly thank our colleagues from INRIA for helping us with our pilot human comparison study: Virginie Mouilleron, Gabrielle Le Bellier, Elodie Etienne, Malik Marmonier, Adle Ben Salem, Célia Nouri, Panagiotis Tsolakis, and Nicolas Angleraud. We also thank Virginie Mouilleron for helping us with the verification of the Llama-generated annotations and Wissam Antoun for verifying the presence of our dataset in FineWeb2 and for helpful discussions of the results.

This work was performed using HPC resources from GENCI–IDRIS (Grant 2026-AD011017684) on the supercomputer Jean Zay’s CSL, A100, and H100 partitions. This project was supported by the BPI Scribe project as well as by Djamé Seddah's PRAIRIE-PSAI chair, funded by the French national agency ANR, as part of the “France 2030” strategy under the reference ANR-23-IACL-0008.

\bibliography{latex/custom}

\appendix

\section{Dataset statistics}
\label{app:dataset_statistics}

This section provides additional statistics for \textsc{TriviaRoomQA-French}. Table~\ref{tab:stat_per_continent} reports the distribution of questions and topics by continent, Table~\ref{tab:stat_per_timeperiod} reports the distribution by time period, and Table~\ref{tab:stat_per_category} reports the distribution by category. Finally, Table~\ref{tab:categories_topics} lists the topics associated with each category. These statistics are used to contextualize the analyses in Section~\ref{sec:results}, especially because continent and time-period results can be affected by the distribution of topics and question difficulty.

\begin{table}[h]
    \centering
    \begin{tabular}{ccc}
    \hline
    \textbf{Continent} & \textbf{\# Questions}  & \textbf{\# Topics} \\
    \hline
Europe  &  4288  &  258 \\
North America  &  1955  &  228 \\
Asia  &  690  &  160 \\
unknown  &  1065  &  121 \\
Africa  &  341  &  116 \\
South America  &  198  &  79 \\
Australia  &  66  &  49 \\
Antarctica  &  35  &  6 \\
Arctic  &  2  &  2 \\
\hline
    \end{tabular}
    \caption{Number of questions and topics per continent in \textsc{TriviaRoomQA-French}.}
    \label{tab:stat_per_continent}
\end{table}

\begin{table}[h]
    \centering
    \begin{tabular}{cc}
    \hline
    \textbf{Time period} & \textbf{\# Questions} \\
    \hline
Timeless & 1742 \\
Historical & 1652 \\
2010s & 1601 \\
2000s & 1277 \\
1990s & 713 \\
1980s & 646 \\
1970s & 316 \\
1960s & 280 \\
1950s & 218 \\
2020s & 195 \\
\hline
    \end{tabular}
    \caption{Number of questions per time period in \textsc{TriviaRoomQA-French}.}
    \label{tab:stat_per_timeperiod}
\end{table}

\begin{table}[h!]
    \centering
    \begin{tabular}{ccc}
    \hline
    \textbf{Category} & \textbf{\# Questions} & \textbf{\# Topics} \\
    \hline
Brands  &  90  &  3 \\
Living creatures  &  330  &  11 \\
Famous places  &  330  &  11 \\
People  &  1440  &  48 \\
Movie  &  1080  &  36 \\
Places  &  480  &  16 \\
Geography  &  240  &  8 \\
Programming  &  360  &  12 \\
Sport  &  630  &  21 \\
Food  &  480  &  16 \\
General knowledge  &  600  &  20 \\
Health  &  120  &  4 \\
Math  &  90  &  3 \\
Nature  &  150  &  5 \\
History  &  180  &  6 \\
Objects  &  150  &  5 \\
Miscellaneous  &  210  &  7 \\
Colors  &  60  &  2 \\
Transport  &  90  &  3 \\
Music  &  210  &  7 \\
News  &  960  &  32 \\
Language  &  210  &  7 \\
Celebrations  &  90  &  3 \\
Chemistry  &  60  &  2 \\
\hline
    \end{tabular}
   \caption{Number of questions and topics per category in \textsc{TriviaRoomQA-French}.}
    \label{tab:stat_per_category}
\end{table}

\section{Data annotation prompts}
\label{app:annotation_prompts}

Tables~\ref{tab:continent-prompt} and~\ref{tab:time-period-prompt} show the prompts used to annotate questions with continent and time period, respectively. The prompts are written in English, while the example questions are provided in French to ensure consistency with the dataset language.

\section{Human validation of the automatic annotations}
\label{app:human_validation}

As the continent and time-period annotations were produced automatically using the Llama-70B-Instruct model, we manually validate a sample of 100 random questions from \textsc{TriviaRoomQA-French} using two annotators. Table~\ref{tab:human_validation} reports the resulting automatic annotation accuracy and inter-annotator agreement. Continent annotations achieve 92\% accuracy, while time-period annotations achieve 79.5\%, with raw inter-annotator agreement of 0.90 and 0.87, respectively. At the same time, the relatively low Cohen's $\kappa$ for continent annotation may partially result from the ambiguity of assigning geographic metadata to some questions, especially those related to fictional events or concerning several countries at the same time. Previous work has already shown that annotator disagreement can arise from uncertainty in item interpretation \citep{jiang-marneffe-2022-investigating}. These results show that the automatically produced metadata can be used as approximate annotations for our aggregate analyses, but we do not use them as gold-standard labels.

\begin{table}[t]
\centering
\begin{adjustbox}{max width=0.48\textwidth}
\begin{tabular}{lcc}
\toprule
\textbf{Metric} & \textbf{Continent} & \textbf{Time period} \\
\midrule
Automatic annotation accuracy & 0.920 & 0.795 \\
Raw inter-annotator agreement & 0.900 & 0.870 \\
Cohen's $\kappa$ & 0.324 & 0.604 \\
Positive agreement & 0.946 & 0.918 \\
Negative agreement & 0.375 & 0.683 \\
\bottomrule
\end{tabular}
\end{adjustbox}
\caption{Human validation of the automatically generated continent and time-period annotations on 100 questions from the \textsc{TriviaRoomQA-French}.}
\label{tab:human_validation}
\end{table}

\section{Human evaluation of the six hardest and easiest \textsc{TriviaRoomQA-French} topics}
\label{app:human_eval}

In this section, we provide details on the pilot experiment comparing average human performance with the performance of the 30 LLMs used in our evaluation.

We selected the six easiest and hardest topics for models by computing average accuracy score for each topic in \textsc{TriviaRoomQA-French}. 

The easiest topics are: \textit{Cities of the world, City nicknames, Alan Turing, Famous Alberts, Microsoft, Internet Frenglish}. 

The hardest topics are: \textit{Cheeses of France, Animals of all kinds, Jean-Marie Bigard, Short quotes, Mike Horn, John McEnroe}\footnote{Jean-Marie Bigard is a French humorist, Mike Horn is a South African–born Swiss explorer, and John McEnroe is an American tennis player.}.

Five French-speaking participants were recruited from among colleagues and were informed about the purpose of the study. All participants took part voluntarily. They answered questions from the twelve topics using the same multiple-choice format as the models.

Participants had an average age of 29.9 years ($\pm$5.6); 60\% were women and 40\% were men. Four participants were French, and one was Belgian.

This experiment is intended as a pilot comparison rather than a definitive estimate of human quiz performance and its goal is to compare general difficulty patterns: whether performance decreases gradually with annotated difficulty, and whether this behavior differs between humans and LLMs.

\section{Model enhancement with search tool}
\label{app:search_tool}

\begin{table}[]
    \centering
    \begin{tabular}{lcc}
    \hline
    \textbf{Model} & \textbf{Base} & \textbf{With search} \\
    \hline
    Qwen3.5-9B & 0.47 & 0.79 \\
    %Aya-Expanse-8B & 0.62 &  \\
    Llama-3.1-8B & 0.51 & 0.63 \\
    \hline 
    \end{tabular}
    \caption{Comparison of model accuracy on \textsc{TriviaRoomQA-French} with and without access to a DuckDuckGo search tool.}
    \label{tab:comparison_agent_noagent}
\end{table}

To test whether web access improves performance on \textsc{TriviaRoomQA-French}, we run an additional experiment with tool-using models under 20B parameters: Qwen3.5-9B and Llama-3.1-8B\footnote{Models are served using Ollama and prompted using the Pydantic AI library.}. We provide them with a DuckDuckGo search tool and compare their performance with and without search access. Table~\ref{tab:comparison_agent_noagent} reports the results.

In this preliminary experiment, search access improves performance for both tested models, with a particularly large gain for Qwen3.5-9B. This suggests that some \textsc{TriviaRoomQA} questions involve information that is difficult to retrieve from model parameters alone. However, the different gains also indicate that the improvement of scores depends on the model’s ability to formulate useful search queries, which might depend on the model's recency and tool-use-specific enhancements, as in the case of Qwen.

\section{Benchmark examples}
\label{app:benchmark_examples}

Table \ref{tab:benchmark_examples} presents examples of questions from different topics in our benchmark, along with their possible answers, correct answer, topic, and difficulty level.

\begin{table*}[]
    \centering
    \begin{adjustbox}{max width=\textwidth}
    \begin{tabular}{p{0.97\textwidth}}
    \hline

    {1.} What did the rich of ancient Greece eat at breakfast? \\
    \hspace{0.5cm}\textbf{Possible answers:} Fish and syrup, Bread and pure wine, Cereals and water, Fruits and honey \\
    \hspace{0.5cm}\textbf{Correct answer:} Bread and pure wine
    \hfill
    \textbf{Topic:} Breakfast
    \hfill
    \textbf{Difficulty:} Confirmed \\
    \hline

    {2.} Which former bodybuilder was governor of California from 2003 to 2011? \\
    \hspace{0.5cm}\textbf{Possible answers:} Reagan, Schwarzenegger, Davis, Rolph \\
    \hspace{0.5cm}\textbf{Correct answer:} Schwarzenegger
    \hfill
    \textbf{Topic:} Mixed general knowledge
    \hfill
    \textbf{Difficulty:} Beginner \\
    \hline

    {3.} Who is one of the three best French football coaches of the 20th century? \\
    \hspace{0.5cm}\textbf{Possible answers:} Albert Dieudonné, Albert Finney, Albert Batteux, Albert Lebrun \\
    \hspace{0.5cm}\textbf{Correct answer:} Albert Batteux
    \hfill
    \textbf{Topic:} Famous Alberts
    \hfill
    \textbf{Difficulty:} Expert \\
    \hline

    {4.} What traditional Japanese art consists of the miniaturizing cultivation of plants? \\
    \hspace{0.5cm}\textbf{Possible answers:} Bankan, Bonsai, Niwaki, Oden \\
    \hspace{0.5cm}\textbf{Correct answer:} Bonsai
    \hfill
    \textbf{Topic:} Japanese folklore
    \hfill
    \textbf{Difficulty:} Beginner \\
    \hline

    {5.} Which industrial cheese is generally coated with paraffin dyed red? \\
    \hspace{0.5cm}\textbf{Possible answers:} Babybel, Saint-paulin, Epoisses, Chavroux \\
    \hspace{0.5cm}\textbf{Correct answer:} Babybel
    \hfill
    \textbf{Topic:} French cheeses
    \hfill
    \textbf{Difficulty:} Beginner \\
    \hline

    \end{tabular}
    \end{adjustbox}
    \caption{Examples of questions from the benchmark across different topics and difficulty levels.}
    \label{tab:benchmark_examples}
\end{table*}

\begin{figure}[ht!]
\centering
\includegraphics[width=\linewidth]{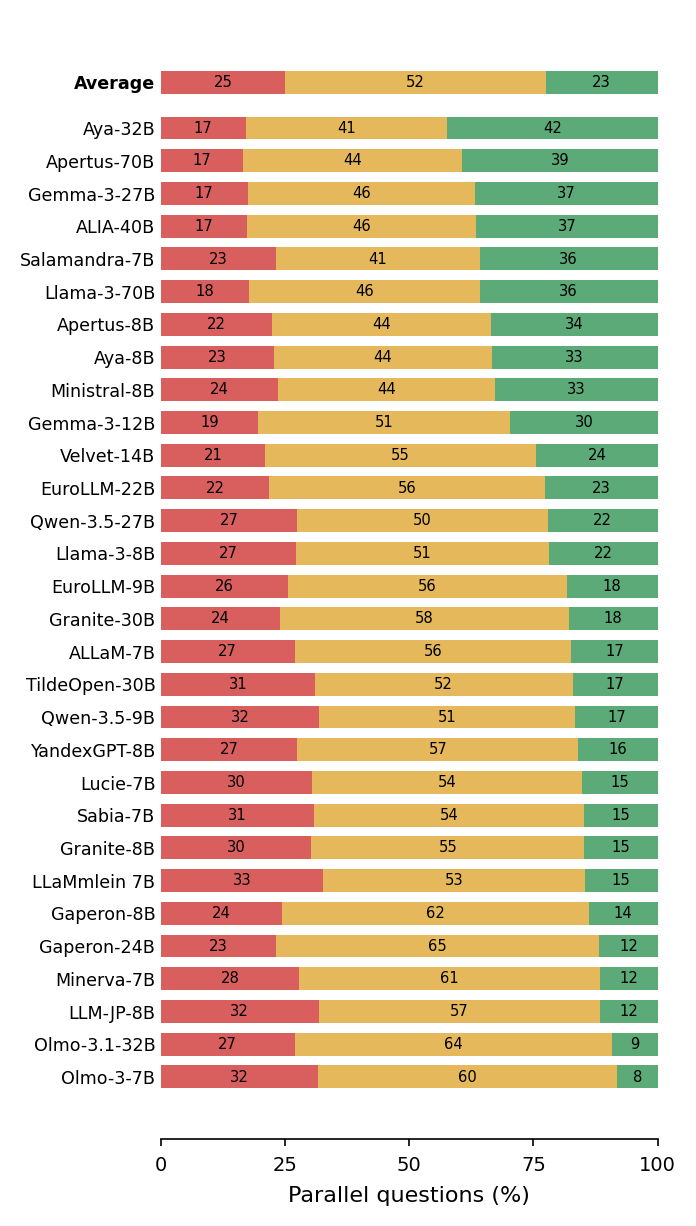}
\caption{Percentage of questions answered correctly by each model in 0 (red), 1-5 (yellow), or 6 (green) languages on the 3,300 parallel questions from \textsc{TriviaRoomQA-Multi}. The average across all models is shown at the top.}
\label{fig:crosslingual_consistency}
\end{figure}

\begin{table}[ht]
\centering
\begin{adjustbox}{max width = 0.9\textwidth}
\begin{tabular}{lc}
\hline
\textbf{Category} & \textbf{Mean accuracy [$\pm$std]} \\
\hline
Music              & 40.8 [$\pm$13.0] \\
Colors             & 41.4 [$\pm$18.8] \\
Food               & 41.5 [$\pm$10.9] \\
Objects            & 45.6 [$\pm$27.9] \\
Living Creatures   & 47.1 [$\pm$15.1] \\
Programming        & 48.6 [$\pm$16.6] \\
People             & 49.6 [$\pm$5.0] \\
Movie              & 49.7 [$\pm$3.7] \\
Celebrations       & 50.1 [$\pm$17.2] \\
Nature             & 51.5 [$\pm$11.5] \\
Places             & 54.4 [$\pm$6.4] \\
Famous Places      & 54.8 [$\pm$19.0] \\
Sport              & 55.8 [$\pm$7.2] \\
Language           & 56.5 [$\pm$21.0] \\
General knowledge  & 56.5 [$\pm$5.8] \\
Geography          & 56.9 [$\pm$8.8] \\
Miscellaneous      & 58.5 [$\pm$6.0] \\
Transport          & 59.0 [$\pm$11.5] \\
Chemistry          & 61.3 [$\pm$4.7] \\
Brands             & 62.3 [$\pm$15.3] \\
News               & 63.8 [$\pm$14.9] \\
Health             & 64.1 [$\pm$30.8] \\
Math               & 68.8 [$\pm$9.7] \\
History            & 75.4 [$\pm$14.7] \\
\hline
\end{tabular}
\end{adjustbox}
\caption{Mean accuracy and standard deviation across time periods for each category for questions from the 1950s to the 2020s in \textsc{TriviaRoomQA-French}. Timeless and historical questions are excluded.}
\label{tab:category_time}
\end{table}

\section{Cross-lingual inconsistency in model answers}
\label{app:crosslingual}

To further analyze how different models perform across languages on our parallel \textsc{TriviaRoomQA-Multi} dataset, we plot in Figure~\ref{fig:crosslingual_consistency} the percentage of questions for which each model answers correctly in all six languages (green), in 1-5 languages (yellow), or in none of the languages (red). As shown in the figure, on average, 52\% of questions show cross-lingual inconsistency, i.e., they are answered correctly in some languages but incorrectly in others.

We additionally examine cases where at least 10 models answer the English version correctly while failing in at least three of the five remaining languages. We identify 181 such questions (5.5\% of the dataset). These questions span a broad range of categories, most frequently General Knowledge (23 questions), Sport (18), Movie (15), People (15), Food (14), and Living Creatures (14). Notably, models also show lower average accuracy on most of these categories, as discussed in Section~\ref{subsec:category_results}, which suggests that model knowledge in these categories may be sensitive to the language in which a question is presented.

\section{Temporal variation across categories}
\label{app:category_time}

As question difficulty may also depend on the time period associated with the knowledge being tested, we examine whether the differences observed across categories in Section~\ref{subsec:category_results} could be related to their temporal distribution. For each category, we compute the mean accuracy across the eight time periods from the 1950s to the 2020s, excluding timeless and historical questions. Table~\ref{tab:category_time} reports the resulting mean accuracy and standard deviation across time periods. A higher standard deviation indicates that model performance within a category varies more strongly depending on the time period of the questions. 

The results show that performance varies relatively little across time periods for categories such as Chemistry, People, General Knowledge, and Movies, suggesting that their difficulty is at least not solely explained by this factor. At the same time, some other categories, such as Health and History, show higher variation across time periods while also achieving high mean accuracy. These results suggest that differences in the temporal distribution of questions do not alone explain the observed category-level differences.

\section{Detailed error analysis}
\label{app:detailed_error_analysis}

In this section, we provide a more detailed analysis of the categories in which the models fail more and less frequently. Across all categories, 362 questions (4.2\% of the dataset) are answered incorrectly by all 30 models. The highest proportions of such questions are found in Programming (7.2\%), Music (6.2\%), Food and Brands (5.6\% each). In contrast, only 0.6\% of History questions are answered incorrectly by all models, and in the Math category there are no questions that are not answered correctly by at least one model.

At the topic level, in the People category, fewer than 10 models answer more than 50\% of the questions correctly for topics such as Maxime Chattam, a French writer, Maria Sharapova, a Russian tennis player, and famous French film directors. The same pattern can be observed in the Food category for questions about Belgian beers, in the Living Creatures category for questions about horses and pigeons, and in the Language category for questions about famous quotations. Examples of questions answered incorrectly by all models include: (i) \textit{What color is the label on a bottle of Delirium Tremens?} and (ii) \textit{Where is the pigeon’s uropygial gland, a gland specific to birds, located?}

All in all, these results suggest that lower model performance on some topics may be associated with several sources of difficulty, including questions about specific named entities and fine-grained factual knowledge that is not necessarily related to a particular named entity.

\section{Statistical Uncertainty of the Results}
\label{app:uncertainty}

We report 95\% confidence intervals (CIs) obtained from 1,000 non-parametric bootstrap resamples of the benchmark questions for the main analyses. For the multilingual analysis, parallel versions of the same underlying question are resampled together. For analyses aggregated across models, we additionally report the standard deviation (std) across models. The following tables provide uncertainty estimates for the language (Table \ref{app:language_ci}), category (Table \ref{app:categories_ci}), difficulty (Table \ref{app:difficulty_ci}), temporal (Table \ref{app:timeperiod_ci}) and geographic (Table \ref{app:continent_ci}) analyses presented in the main text.

\begin{table*}
\begin{adjustbox}{max width = \textwidth}
\begin{tabular}{lcccccc}
\hline
 & \textbf{English} & \textbf{French} & \textbf{Italian} & \textbf{German} & \textbf{Spanish} & \textbf{Dutch} \\
\hline
Aya-32B & 0.67 [0.65, 0.68] & 0.69 [0.67, 0.70] & 0.62 [0.60, 0.63] & 0.62 [0.61, 0.64] & 0.61 [0.59, 0.63] & 0.61 [0.60, 0.63] \\
Apertus-70B & 0.63 [0.62, 0.65] & 0.67 [0.65, 0.68] & 0.60 [0.59, 0.62] & 0.62 [0.60, 0.64] & 0.62 [0.60, 0.64] & 0.60 [0.58, 0.62] \\
Gemma-3-27B & 0.63 [0.62, 0.65] & 0.66 [0.64, 0.68] & 0.60 [0.58, 0.61] & 0.59 [0.57, 0.60] & 0.59 [0.57, 0.60] & 0.59 [0.57, 0.61] \\
ALIA-40B & 0.61 [0.60, 0.63] & 0.65 [0.64, 0.67] & 0.59 [0.57, 0.60] & 0.59 [0.57, 0.61] & 0.59 [0.58, 0.61] & 0.59 [0.57, 0.60] \\
Llama-3-70B & 0.63 [0.61, 0.65] & 0.66 [0.64, 0.68] & 0.58 [0.56, 0.60] & 0.58 [0.56, 0.60] & 0.58 [0.56, 0.59] & 0.58 [0.56, 0.59] \\
Salamandra-7B & 0.59 [0.57, 0.61] & 0.62 [0.60, 0.64] & 0.57 [0.55, 0.58] & 0.55 [0.53, 0.57] & 0.55 [0.54, 0.57] & 0.55 [0.53, 0.56] \\
Apertus-8B & 0.57 [0.55, 0.59] & 0.58 [0.57, 0.60] & 0.55 [0.53, 0.57] & 0.55 [0.53, 0.57] & 0.56 [0.54, 0.57] & 0.54 [0.53, 0.56] \\
Ministral-8B & 0.60 [0.59, 0.62] & 0.62 [0.60, 0.63] & 0.55 [0.54, 0.57] & 0.53 [0.51, 0.54] & 0.54 [0.53, 0.56] & 0.50 [0.48, 0.51] \\
Gemma-3-12B & 0.59 [0.58, 0.61] & 0.61 [0.59, 0.62] & 0.53 [0.51, 0.55] & 0.54 [0.53, 0.56] & 0.52 [0.51, 0.54] & 0.53 [0.51, 0.55] \\
Aya-8B & 0.58 [0.56, 0.60] & 0.60 [0.58, 0.61] & 0.54 [0.52, 0.55] & 0.54 [0.53, 0.56] & 0.54 [0.52, 0.55] & 0.53 [0.51, 0.55] \\
Velvet-14B & 0.56 [0.55, 0.58] & 0.57 [0.56, 0.59] & 0.54 [0.53, 0.56] & 0.51 [0.49, 0.52] & 0.53 [0.52, 0.55] & 0.43 [0.41, 0.45] \\
EuroLLM-22B & 0.51 [0.49, 0.52] & 0.55 [0.53, 0.56] & 0.49 [0.47, 0.51] & 0.47 [0.45, 0.49] & 0.50 [0.49, 0.52] & 0.46 [0.44, 0.47] \\
Qwen-3.5-27B & 0.50 [0.48, 0.52] & 0.50 [0.48, 0.52] & 0.46 [0.45, 0.48] & 0.46 [0.45, 0.48] & 0.45 [0.43, 0.47] & 0.45 [0.43, 0.46] \\
Llama-3-8B & 0.52 [0.50, 0.53] & 0.49 [0.47, 0.50] & 0.46 [0.44, 0.48] & 0.45 [0.43, 0.46] & 0.44 [0.42, 0.46] & 0.45 [0.44, 0.47] \\
Granite-30B & 0.53 [0.51, 0.54] & 0.49 [0.48, 0.51] & 0.43 [0.41, 0.45] & 0.44 [0.42, 0.46] & 0.44 [0.42, 0.46] & 0.44 [0.42, 0.46] \\
EuroLLM-9B & 0.46 [0.44, 0.48] & 0.47 [0.46, 0.49] & 0.43 [0.42, 0.45] & 0.42 [0.41, 0.44] & 0.45 [0.43, 0.47] & 0.43 [0.41, 0.44] \\
ALLaM-7B & 0.51 [0.49, 0.52] & 0.46 [0.44, 0.48] & 0.43 [0.41, 0.45] & 0.42 [0.40, 0.44] & 0.40 [0.39, 0.42] & 0.41 [0.39, 0.43] \\
YandexGPT-8B & 0.49 [0.47, 0.50] & 0.43 [0.41, 0.45] & 0.40 [0.39, 0.42] & 0.40 [0.39, 0.42] & 0.42 [0.40, 0.43] & 0.41 [0.39, 0.43] \\
Gaperon-8B & 0.44 [0.42, 0.46] & 0.49 [0.47, 0.50] & 0.40 [0.39, 0.42] & 0.40 [0.38, 0.42] & 0.38 [0.37, 0.40] & 0.41 [0.39, 0.43] \\
Lucie-7B & 0.45 [0.43, 0.46] & 0.47 [0.46, 0.49] & 0.40 [0.39, 0.42] & 0.40 [0.39, 0.42] & 0.42 [0.41, 0.44] & 0.33 [0.32, 0.35] \\
TildeOpen-30B & 0.40 [0.39, 0.42] & 0.44 [0.42, 0.45] & 0.41 [0.39, 0.43] & 0.41 [0.40, 0.43] & 0.40 [0.39, 0.42] & 0.41 [0.39, 0.43] \\
Qwen-3.5-9B & 0.43 [0.42, 0.45] & 0.43 [0.42, 0.45] & 0.40 [0.38, 0.41] & 0.40 [0.39, 0.42] & 0.39 [0.37, 0.41] & 0.39 [0.38, 0.41] \\
Gaperon-24B & 0.45 [0.43, 0.47] & 0.47 [0.45, 0.48] & 0.39 [0.37, 0.40] & 0.38 [0.37, 0.40] & 0.38 [0.36, 0.39] & 0.38 [0.37, 0.40] \\
Granite-8B & 0.46 [0.44, 0.48] & 0.40 [0.38, 0.41] & 0.38 [0.37, 0.40] & 0.39 [0.37, 0.40] & 0.39 [0.37, 0.40] & 0.39 [0.38, 0.41] \\
Sabia-7B & 0.45 [0.43, 0.46] & 0.41 [0.40, 0.43] & 0.38 [0.36, 0.40] & 0.38 [0.36, 0.40] & 0.40 [0.38, 0.41] & 0.38 [0.36, 0.40] \\
LLaMmlein-7B & 0.43 [0.41, 0.45] & 0.39 [0.37, 0.40] & 0.36 [0.34, 0.37] & 0.46 [0.44, 0.48] & 0.36 [0.34, 0.37] & 0.37 [0.36, 0.39] \\
Minerva-7B & 0.45 [0.44, 0.47] & 0.38 [0.36, 0.39] & 0.44 [0.43, 0.46] & 0.34 [0.32, 0.36] & 0.37 [0.36, 0.39] & 0.34 [0.33, 0.36] \\
Olmo-3.1-32B & 0.45 [0.43, 0.47] & 0.39 [0.38, 0.41] & 0.37 [0.35, 0.39] & 0.37 [0.35, 0.38] & 0.33 [0.31, 0.34] & 0.36 [0.34, 0.37] \\
LLM-JP-8B & 0.42 [0.41, 0.44] & 0.37 [0.35, 0.39] & 0.35 [0.34, 0.37] & 0.36 [0.35, 0.38] & 0.36 [0.34, 0.38] & 0.36 [0.35, 0.38] \\
Olmo-3-7B & 0.38 [0.36, 0.40] & 0.34 [0.32, 0.36] & 0.32 [0.30, 0.34] & 0.32 [0.30, 0.33] & 0.32 [0.30, 0.33] & 0.32 [0.30, 0.33] \\
\hline
\end{tabular}
\end{adjustbox}
\caption{Accuracy and 95\% bootstrap confidence intervals for each model across languages reported in Figure \ref{fig:multiling_res}.}
\label{app:language_ci}
\end{table*}

\begin{table}
\begin{adjustbox}{max width = 0.48\textwidth}
\begin{tabular}{lrrr}
\hline
\textbf{Category} & \textbf{Size} & \textbf{Accuracy [$\pm$std]} & \textbf{95\% CI} \\
\hline
Music & 210 & 0.46 [$\pm$ 0.10] & [0.41, 0.50] \\
Nature & 150 & 0.47 [$\pm$ 0.10] & [0.43, 0.52] \\
Food & 480 & 0.47 [$\pm$ 0.10] & [0.44, 0.50] \\
Colors & 60 & 0.49 [$\pm$ 0.14] & [0.41, 0.57] \\
Movie & 1080 & 0.49 [$\pm$ 0.11] & [0.48, 0.51] \\
People & 1440 & 0.50 [$\pm$ 0.10] & [0.48, 0.51] \\
News & 960 & 0.50 [$\pm$ 0.10] & [0.48, 0.52] \\
Living creatures & 330 & 0.51 [$\pm$ 0.11] & [0.47, 0.54] \\
Sport & 630 & 0.52 [$\pm$ 0.11] & [0.50, 0.54] \\
Miscellaneous & 210 & 0.55 [$\pm$ 0.12] & [0.51, 0.59] \\
Celebrations & 90 & 0.56 [$\pm$ 0.14] & [0.50, 0.62] \\
Transport & 90 & 0.57 [$\pm$ 0.14] & [0.50, 0.62] \\
Programming & 360 & 0.57 [$\pm$ 0.10] & [0.53, 0.60] \\
Brands & 90 & 0.57 [$\pm$ 0.10] & [0.50, 0.64] \\
Famous places & 330 & 0.58 [$\pm$ 0.13] & [0.54, 0.61] \\
General knowledge & 600 & 0.58 [$\pm$ 0.13] & [0.56, 0.60] \\
Chemistry & 60 & 0.58 [$\pm$ 0.10] & [0.49, 0.67] \\
Language & 210 & 0.59 [$\pm$ 0.09] & [0.55, 0.64] \\
Places & 480 & 0.60 [$\pm$ 0.11] & [0.58, 0.63] \\
Health & 120 & 0.61 [$\pm$ 0.11] & [0.56, 0.66] \\
Objects & 150 & 0.61 [$\pm$ 0.13] & [0.56, 0.66] \\
Geography & 240 & 0.63 [$\pm$ 0.11] & [0.59, 0.67] \\
Math & 90 & 0.67 [$\pm$ 0.17] & [0.62, 0.72] \\
History & 180 & 0.69 [$\pm$ 0.13] & [0.65, 0.73] \\
\hline
\end{tabular}
\end{adjustbox}
\caption{Mean accuracy, standard deviation, and 95\% bootstrap confidence intervals for each category across models, corresponding to Figure~\ref{fig:category_results}.}

\label{app:categories_ci}
\end{table}

\begin{table}
\begin{adjustbox}{max width = 0.48\textwidth}
\begin{tabular}{lrrr}
\hline
\textbf{Difficulty} & \textbf{Size} & \textbf{Accuracy [$\pm$std]} & \textbf{95\% CI} \\
\hline
Beginner & 2900 & 0.60 [$\pm$ 0.12] & [0.59, 0.61] \\
Confirmed & 2870 & 0.52 [$\pm$ 0.11] & [0.51, 0.53] \\
Expert & 2870 & 0.48 [$\pm$ 0.10] & [0.47, 0.49] \\
\hline
\end{tabular}
\end{adjustbox}
\caption{Mean accuracy, standard deviation across models, and 95\% bootstrap confidence intervals for each difficulty level reported in Figure \ref{fig:detailed_niveau_perf}.}
\label{app:difficulty_ci}
\end{table}

\begin{table*}
\begin{adjustbox}{max width = \textwidth}

\begin{tabular}{lcccccccccc}
\hline
 & \textbf{Timeless} & \textbf{Historical} & \textbf{1950s} & \textbf{1960s} & \textbf{1970s} & \textbf{1980s} & \textbf{1990s} & \textbf{2000s} & \textbf{2010s} & \textbf{2020s} \\
& (1742 q.) & (1652 q.) & (218 q.) & (280 q.) & (316 q.) & (646 q.) & (713 q.) & (1277 q.) & (1601 q.) & (195 q.) \\
\hline
Velvet-14B & 0.61 [0.59, 0.63] & 0.68 [0.66, 0.71] & 0.63 [0.57, 0.70] & 0.60 [0.55, 0.66] & 0.60 [0.54, 0.66] & 0.55 [0.51, 0.59] & 0.54 [0.50, 0.57] & 0.57 [0.55, 0.60] & 0.57 [0.55, 0.60] & 0.59 [0.53, 0.66] \\
Apertus-8B & 0.61 [0.59, 0.63] & 0.73 [0.71, 0.75] & 0.67 [0.61, 0.72] & 0.64 [0.59, 0.70] & 0.64 [0.58, 0.69] & 0.57 [0.54, 0.61] & 0.60 [0.56, 0.64] & 0.59 [0.57, 0.62] & 0.55 [0.53, 0.58] & 0.63 [0.56, 0.70] \\
Aya-8B & 0.62 [0.60, 0.65] & 0.69 [0.67, 0.71] & 0.63 [0.57, 0.70] & 0.61 [0.55, 0.66] & 0.60 [0.54, 0.65] & 0.54 [0.50, 0.58] & 0.57 [0.54, 0.61] & 0.62 [0.59, 0.64] & 0.59 [0.57, 0.62] & 0.58 [0.51, 0.65] \\
EuroLLM-9B & 0.53 [0.51, 0.56] & 0.62 [0.59, 0.64] & 0.61 [0.54, 0.67] & 0.49 [0.43, 0.55] & 0.49 [0.44, 0.55] & 0.42 [0.39, 0.46] & 0.47 [0.44, 0.51] & 0.48 [0.45, 0.51] & 0.42 [0.40, 0.45] & 0.45 [0.38, 0.52] \\
Gaperon-8B & 0.55 [0.52, 0.57] & 0.59 [0.57, 0.62] & 0.58 [0.52, 0.65] & 0.52 [0.46, 0.57] & 0.49 [0.44, 0.54] & 0.44 [0.41, 0.48] & 0.48 [0.44, 0.51] & 0.48 [0.46, 0.51] & 0.46 [0.44, 0.48] & 0.47 [0.39, 0.53] \\
Ministral-8B & 0.65 [0.62, 0.67] & 0.73 [0.71, 0.75] & 0.68 [0.62, 0.74] & 0.64 [0.59, 0.70] & 0.63 [0.58, 0.69] & 0.58 [0.54, 0.62] & 0.63 [0.59, 0.66] & 0.62 [0.59, 0.65] & 0.60 [0.57, 0.62] & 0.63 [0.55, 0.69] \\
Llama-3-8B & 0.49 [0.47, 0.52] & 0.59 [0.57, 0.61] & 0.60 [0.54, 0.67] & 0.48 [0.42, 0.54] & 0.53 [0.48, 0.59] & 0.49 [0.44, 0.52] & 0.50 [0.46, 0.54] & 0.50 [0.47, 0.52] & 0.47 [0.45, 0.49] & 0.45 [0.38, 0.52] \\
LLaMmlein 7B & 0.36 [0.34, 0.38] & 0.43 [0.41, 0.46] & 0.41 [0.34, 0.47] & 0.38 [0.32, 0.44] & 0.37 [0.31, 0.42] & 0.39 [0.36, 0.43] & 0.37 [0.33, 0.40] & 0.41 [0.38, 0.43] & 0.38 [0.35, 0.40] & 0.37 [0.30, 0.44] \\
Lucie-7B & 0.51 [0.48, 0.53] & 0.58 [0.56, 0.61] & 0.52 [0.46, 0.59] & 0.47 [0.41, 0.53] & 0.47 [0.42, 0.53] & 0.42 [0.38, 0.46] & 0.46 [0.42, 0.49] & 0.47 [0.44, 0.49] & 0.45 [0.43, 0.48] & 0.52 [0.44, 0.58] \\
Olmo-3-7B & 0.37 [0.35, 0.39] & 0.38 [0.35, 0.40] & 0.39 [0.32, 0.45] & 0.38 [0.33, 0.44] & 0.35 [0.30, 0.41] & 0.33 [0.29, 0.37] & 0.34 [0.30, 0.37] & 0.36 [0.33, 0.38] & 0.33 [0.30, 0.35] & 0.39 [0.32, 0.46] \\
Qwen-3.5-9B & 0.49 [0.47, 0.52] & 0.56 [0.54, 0.59] & 0.52 [0.45, 0.59] & 0.49 [0.43, 0.54] & 0.51 [0.46, 0.57] & 0.42 [0.38, 0.46] & 0.43 [0.40, 0.47] & 0.44 [0.41, 0.46] & 0.40 [0.38, 0.42] & 0.45 [0.38, 0.52] \\
Gemma-3-12B & 0.65 [0.63, 0.67] & 0.71 [0.69, 0.73] & 0.67 [0.61, 0.73] & 0.64 [0.58, 0.69] & 0.63 [0.59, 0.69] & 0.58 [0.54, 0.61] & 0.61 [0.57, 0.65] & 0.63 [0.60, 0.65] & 0.59 [0.57, 0.62] & 0.61 [0.54, 0.67] \\
ALLaM-7B & 0.47 [0.45, 0.49] & 0.55 [0.53, 0.58] & 0.53 [0.46, 0.59] & 0.51 [0.46, 0.57] & 0.55 [0.50, 0.61] & 0.44 [0.40, 0.48] & 0.49 [0.46, 0.53] & 0.49 [0.46, 0.52] & 0.43 [0.41, 0.46] & 0.43 [0.35, 0.49] \\
Granite-8B & 0.44 [0.42, 0.47] & 0.49 [0.46, 0.51] & 0.46 [0.39, 0.53] & 0.41 [0.35, 0.46] & 0.47 [0.41, 0.52] & 0.40 [0.37, 0.44] & 0.38 [0.35, 0.42] & 0.42 [0.40, 0.45] & 0.38 [0.35, 0.40] & 0.36 [0.29, 0.43] \\
LLM-JP-8B & 0.39 [0.37, 0.42] & 0.44 [0.42, 0.47] & 0.46 [0.39, 0.52] & 0.41 [0.35, 0.47] & 0.39 [0.33, 0.44] & 0.39 [0.35, 0.43] & 0.38 [0.34, 0.42] & 0.39 [0.37, 0.42] & 0.35 [0.33, 0.38] & 0.35 [0.29, 0.42] \\
Sabia-7B & 0.41 [0.39, 0.44] & 0.49 [0.46, 0.51] & 0.49 [0.42, 0.56] & 0.44 [0.38, 0.50] & 0.43 [0.37, 0.48] & 0.41 [0.37, 0.45] & 0.43 [0.39, 0.47] & 0.43 [0.40, 0.46] & 0.41 [0.39, 0.43] & 0.39 [0.32, 0.46] \\
Salamandra-7B & 0.63 [0.60, 0.65] & 0.73 [0.71, 0.75] & 0.69 [0.63, 0.76] & 0.63 [0.57, 0.69] & 0.63 [0.58, 0.68] & 0.58 [0.54, 0.62] & 0.61 [0.57, 0.64] & 0.60 [0.57, 0.63] & 0.58 [0.56, 0.61] & 0.67 [0.61, 0.73] \\
Minerva-7B & 0.36 [0.34, 0.38] & 0.42 [0.40, 0.45] & 0.43 [0.37, 0.50] & 0.41 [0.35, 0.47] & 0.42 [0.37, 0.48] & 0.39 [0.35, 0.43] & 0.38 [0.35, 0.42] & 0.38 [0.36, 0.41] & 0.38 [0.36, 0.40] & 0.39 [0.32, 0.46] \\
YandexGPT-8B & 0.46 [0.44, 0.48] & 0.52 [0.49, 0.54] & 0.47 [0.40, 0.54] & 0.44 [0.38, 0.49] & 0.45 [0.40, 0.51] & 0.38 [0.35, 0.42] & 0.42 [0.38, 0.46] & 0.44 [0.42, 0.47] & 0.41 [0.39, 0.43] & 0.45 [0.37, 0.52] \\
Apertus-70B & 0.68 [0.66, 0.70] & 0.77 [0.75, 0.79] & 0.79 [0.73, 0.84] & 0.71 [0.65, 0.76] & 0.71 [0.66, 0.76] & 0.65 [0.61, 0.68] & 0.67 [0.63, 0.70] & 0.70 [0.68, 0.73] & 0.66 [0.64, 0.69] & 0.68 [0.61, 0.75] \\
ALIA-40B & 0.65 [0.63, 0.67] & 0.77 [0.75, 0.79] & 0.71 [0.65, 0.77] & 0.71 [0.66, 0.77] & 0.73 [0.68, 0.78] & 0.63 [0.59, 0.66] & 0.67 [0.64, 0.70] & 0.69 [0.67, 0.71] & 0.65 [0.62, 0.67] & 0.65 [0.58, 0.72] \\
Aya-32B & 0.70 [0.67, 0.72] & 0.79 [0.77, 0.81] & 0.79 [0.74, 0.85] & 0.74 [0.69, 0.79] & 0.73 [0.68, 0.78] & 0.68 [0.65, 0.72] & 0.68 [0.65, 0.71] & 0.72 [0.70, 0.74] & 0.71 [0.69, 0.73] & 0.66 [0.59, 0.72] \\
EuroLLM-22B & 0.59 [0.57, 0.61] & 0.65 [0.63, 0.67] & 0.63 [0.57, 0.69] & 0.59 [0.53, 0.64] & 0.58 [0.52, 0.63] & 0.51 [0.47, 0.55] & 0.52 [0.48, 0.55] & 0.55 [0.52, 0.58] & 0.52 [0.49, 0.54] & 0.60 [0.53, 0.67] \\
Llama-3-70B & 0.67 [0.65, 0.69] & 0.77 [0.75, 0.79] & 0.74 [0.68, 0.79] & 0.71 [0.66, 0.76] & 0.65 [0.59, 0.70] & 0.67 [0.63, 0.71] & 0.72 [0.68, 0.75] & 0.68 [0.65, 0.70] & 0.62 [0.60, 0.65] & 0.63 [0.56, 0.69] \\
Qwen-3.5-27B & 0.55 [0.53, 0.58] & 0.65 [0.63, 0.67] & 0.58 [0.52, 0.64] & 0.51 [0.46, 0.57] & 0.49 [0.44, 0.54] & 0.44 [0.41, 0.48] & 0.45 [0.41, 0.49] & 0.49 [0.46, 0.51] & 0.47 [0.44, 0.49] & 0.48 [0.41, 0.55] \\
TildeOpen-30B & 0.47 [0.45, 0.50] & 0.50 [0.47, 0.52] & 0.47 [0.41, 0.54] & 0.44 [0.38, 0.50] & 0.44 [0.39, 0.49] & 0.40 [0.36, 0.44] & 0.42 [0.38, 0.45] & 0.45 [0.42, 0.48] & 0.43 [0.41, 0.46] & 0.51 [0.43, 0.58] \\
Olmo-3.1-32B & 0.40 [0.37, 0.42] & 0.45 [0.43, 0.47] & 0.46 [0.39, 0.52] & 0.43 [0.37, 0.48] & 0.44 [0.39, 0.49] & 0.39 [0.36, 0.43] & 0.40 [0.36, 0.43] & 0.42 [0.39, 0.44] & 0.39 [0.36, 0.41] & 0.35 [0.29, 0.43] \\
Gaperon-24B & 0.51 [0.48, 0.53] & 0.57 [0.55, 0.59] & 0.54 [0.48, 0.61] & 0.50 [0.45, 0.56] & 0.51 [0.45, 0.56] & 0.40 [0.36, 0.44] & 0.41 [0.38, 0.45] & 0.46 [0.43, 0.48] & 0.44 [0.42, 0.47] & 0.48 [0.41, 0.55] \\
Gemma-3-27B & 0.70 [0.67, 0.72] & 0.78 [0.76, 0.80] & 0.74 [0.68, 0.79] & 0.66 [0.61, 0.72] & 0.66 [0.61, 0.72] & 0.62 [0.58, 0.65] & 0.66 [0.63, 0.70] & 0.66 [0.63, 0.68] & 0.65 [0.62, 0.67] & 0.68 [0.62, 0.74] \\
Granite-30B & 0.53 [0.51, 0.55] & 0.58 [0.56, 0.60] & 0.58 [0.51, 0.64] & 0.49 [0.43, 0.55] & 0.51 [0.45, 0.56] & 0.48 [0.44, 0.52] & 0.51 [0.47, 0.54] & 0.52 [0.49, 0.54] & 0.45 [0.43, 0.47] & 0.49 [0.42, 0.56] \\
\hline
\end{tabular}
\end{adjustbox}
\caption{Accuracy and 95\% bootstrap confidence intervals for each model across time periods corresponding to Figures \ref{fig:timeperiod_perf_merged} and \ref{fig:timeperiod_perf}.}
\label{app:timeperiod_ci}
\end{table*}

\begin{table*}
\begin{adjustbox}{max width = \textwidth}

\begin{tabular}{lccccccccc}
\hline
 & \textbf{Africa} & \textbf{Asia} & \textbf{Europe} & \textbf{N. America} & \textbf{S. America} & \textbf{Australia} & \textbf{Antarctica} & \textbf{Arctic} & \textbf{unknown} \\
 & (341 q.) & (690 q.) & (4288 q.) & (1955 q.) & (198 q.) & (66 q.) & (35 q.) & (2 q.) & (1065 q.) \\
\hline
Velvet-14B & 0.66 [0.61, 0.71] & 0.63 [0.59, 0.66] & 0.59 [0.58, 0.60] & 0.62 [0.59, 0.64] & 0.64 [0.58, 0.71] & 0.67 [0.55, 0.77] & 0.57 [0.40, 0.71] & 0.50 [0.00, 1.00] & 0.58 [0.55, 0.60] \\
Apertus-8B & 0.68 [0.63, 0.73] & 0.66 [0.62, 0.70] & 0.59 [0.58, 0.60] & 0.67 [0.65, 0.69] & 0.67 [0.60, 0.73] & 0.59 [0.47, 0.70] & 0.66 [0.51, 0.80] & 0.50 [0.00, 1.00] & 0.59 [0.56, 0.62] \\
Aya-8B & 0.70 [0.65, 0.75] & 0.66 [0.62, 0.69] & 0.58 [0.57, 0.60] & 0.66 [0.64, 0.68] & 0.66 [0.59, 0.73] & 0.67 [0.55, 0.77] & 0.71 [0.57, 0.86] & 0.50 [0.00, 1.00] & 0.61 [0.58, 0.64] \\
EuroLLM-9B & 0.56 [0.51, 0.61] & 0.54 [0.50, 0.58] & 0.49 [0.48, 0.51] & 0.50 [0.48, 0.52] & 0.52 [0.44, 0.58] & 0.56 [0.45, 0.68] & 0.51 [0.37, 0.66] & 0.50 [0.00, 1.00] & 0.51 [0.48, 0.54] \\
Gaperon-8B & 0.56 [0.51, 0.61] & 0.53 [0.50, 0.57] & 0.51 [0.49, 0.52] & 0.50 [0.48, 0.52] & 0.59 [0.52, 0.65] & 0.58 [0.45, 0.68] & 0.69 [0.54, 0.83] & 0.50 [0.00, 1.00] & 0.50 [0.47, 0.53] \\
Ministral-8B & 0.72 [0.67, 0.77] & 0.70 [0.66, 0.73] & 0.61 [0.59, 0.62] & 0.69 [0.67, 0.71] & 0.65 [0.59, 0.72] & 0.61 [0.48, 0.73] & 0.66 [0.51, 0.80] & 0.50 [0.00, 1.00] & 0.64 [0.61, 0.67] \\
Llama-3-8B & 0.57 [0.52, 0.63] & 0.54 [0.51, 0.58] & 0.48 [0.47, 0.50] & 0.57 [0.54, 0.59] & 0.55 [0.47, 0.62] & 0.58 [0.45, 0.70] & 0.60 [0.43, 0.74] & 0.50 [0.00, 1.00] & 0.46 [0.43, 0.49] \\
LLaMmlein 7B & 0.39 [0.34, 0.44] & 0.38 [0.35, 0.42] & 0.38 [0.37, 0.39] & 0.44 [0.41, 0.46] & 0.41 [0.34, 0.47] & 0.45 [0.33, 0.58] & 0.37 [0.23, 0.54] & 0.00 [0.00, 0.00] & 0.34 [0.32, 0.37] \\
Lucie-7B & 0.52 [0.47, 0.57] & 0.50 [0.46, 0.54] & 0.50 [0.48, 0.51] & 0.48 [0.46, 0.51] & 0.50 [0.43, 0.57] & 0.44 [0.32, 0.56] & 0.57 [0.40, 0.71] & 0.50 [0.00, 1.00] & 0.48 [0.45, 0.51] \\
Olmo-3-7B & 0.39 [0.35, 0.45] & 0.37 [0.33, 0.40] & 0.33 [0.32, 0.34] & 0.39 [0.37, 0.42] & 0.36 [0.30, 0.43] & 0.41 [0.30, 0.53] & 0.34 [0.20, 0.51] & 0.50 [0.00, 1.00] & 0.37 [0.34, 0.40] \\
Qwen-3.5-9B & 0.52 [0.47, 0.58] & 0.49 [0.46, 0.53] & 0.44 [0.43, 0.46] & 0.50 [0.48, 0.52] & 0.47 [0.41, 0.54] & 0.45 [0.33, 0.58] & 0.51 [0.37, 0.69] & 0.50 [0.00, 1.00] & 0.50 [0.47, 0.53] \\
Gemma-3-12B & 0.70 [0.65, 0.75] & 0.69 [0.66, 0.73] & 0.60 [0.58, 0.61] & 0.70 [0.68, 0.72] & 0.66 [0.60, 0.73] & 0.64 [0.52, 0.74] & 0.71 [0.57, 0.86] & 0.50 [0.00, 1.00] & 0.63 [0.60, 0.66] \\
ALLaM-7B & 0.53 [0.48, 0.59] & 0.51 [0.47, 0.54] & 0.45 [0.44, 0.47] & 0.55 [0.53, 0.58] & 0.50 [0.43, 0.57] & 0.50 [0.38, 0.62] & 0.46 [0.29, 0.63] & 0.50 [0.00, 1.00] & 0.46 [0.43, 0.49] \\
Granite-8B & 0.45 [0.39, 0.50] & 0.42 [0.39, 0.46] & 0.40 [0.39, 0.42] & 0.48 [0.46, 0.50] & 0.45 [0.38, 0.52] & 0.44 [0.32, 0.56] & 0.37 [0.23, 0.54] & 0.00 [0.00, 0.00] & 0.43 [0.40, 0.46] \\
LLM-JP-8B & 0.43 [0.38, 0.48] & 0.40 [0.37, 0.44] & 0.37 [0.36, 0.39] & 0.44 [0.42, 0.46] & 0.41 [0.34, 0.48] & 0.41 [0.29, 0.53] & 0.31 [0.17, 0.49] & 0.50 [0.00, 1.00] & 0.39 [0.36, 0.42] \\
Sabia-7B & 0.44 [0.39, 0.50] & 0.43 [0.40, 0.47] & 0.42 [0.40, 0.43] & 0.48 [0.45, 0.50] & 0.46 [0.39, 0.53] & 0.44 [0.32, 0.56] & 0.46 [0.29, 0.63] & 0.50 [0.00, 1.00] & 0.42 [0.38, 0.45] \\
Salamandra-7B & 0.69 [0.64, 0.74] & 0.67 [0.63, 0.70] & 0.60 [0.59, 0.62] & 0.67 [0.65, 0.69] & 0.67 [0.60, 0.73] & 0.73 [0.62, 0.83] & 0.63 [0.46, 0.77] & 0.50 [0.00, 1.00] & 0.62 [0.59, 0.65] \\
Minerva-7B & 0.38 [0.33, 0.43] & 0.38 [0.34, 0.42] & 0.39 [0.37, 0.40] & 0.43 [0.41, 0.45] & 0.37 [0.31, 0.44] & 0.32 [0.21, 0.42] & 0.43 [0.26, 0.60] & 0.50 [0.00, 1.00] & 0.34 [0.31, 0.37] \\
YandexGPT-8B & 0.49 [0.44, 0.55] & 0.48 [0.44, 0.51] & 0.42 [0.41, 0.44] & 0.47 [0.45, 0.50] & 0.49 [0.42, 0.56] & 0.47 [0.36, 0.58] & 0.49 [0.31, 0.63] & 0.50 [0.00, 1.00] & 0.47 [0.44, 0.50] \\
Apertus-70B & 0.77 [0.72, 0.81] & 0.69 [0.66, 0.73] & 0.67 [0.66, 0.69] & 0.74 [0.72, 0.76] & 0.76 [0.70, 0.82] & 0.64 [0.52, 0.74] & 0.66 [0.51, 0.80] & 1.00 [1.00, 1.00] & 0.69 [0.66, 0.72] \\
ALIA-40B & 0.76 [0.72, 0.81] & 0.72 [0.68, 0.75] & 0.67 [0.65, 0.68] & 0.72 [0.70, 0.74] & 0.70 [0.64, 0.76] & 0.71 [0.59, 0.82] & 0.71 [0.54, 0.86] & 0.50 [0.00, 1.00] & 0.62 [0.59, 0.65] \\
Aya-32B & 0.79 [0.74, 0.83] & 0.75 [0.72, 0.79] & 0.70 [0.69, 0.72] & 0.76 [0.74, 0.78] & 0.74 [0.68, 0.80] & 0.71 [0.59, 0.82] & 0.69 [0.51, 0.83] & 0.50 [0.00, 1.00] & 0.68 [0.65, 0.70] \\
EuroLLM-22B & 0.59 [0.54, 0.64] & 0.59 [0.55, 0.63] & 0.55 [0.54, 0.57] & 0.59 [0.56, 0.61] & 0.66 [0.60, 0.72] & 0.59 [0.48, 0.70] & 0.57 [0.40, 0.71] & 0.50 [0.00, 1.00] & 0.58 [0.55, 0.61] \\
Llama-3-70B & 0.72 [0.67, 0.77] & 0.72 [0.68, 0.75] & 0.66 [0.64, 0.67] & 0.74 [0.72, 0.76] & 0.69 [0.63, 0.76] & 0.74 [0.64, 0.83] & 0.74 [0.60, 0.89] & 0.50 [0.00, 1.00] & 0.67 [0.64, 0.70] \\
Qwen-3.5-27B & 0.58 [0.53, 0.63] & 0.57 [0.53, 0.61] & 0.50 [0.48, 0.51] & 0.54 [0.52, 0.56] & 0.52 [0.44, 0.59] & 0.53 [0.41, 0.64] & 0.60 [0.43, 0.77] & 0.50 [0.00, 1.00] & 0.55 [0.52, 0.58] \\
TildeOpen-30B & 0.47 [0.42, 0.52] & 0.49 [0.45, 0.53] & 0.45 [0.43, 0.46] & 0.47 [0.44, 0.49] & 0.41 [0.34, 0.48] & 0.45 [0.35, 0.58] & 0.43 [0.26, 0.60] & 0.50 [0.00, 1.00] & 0.46 [0.43, 0.49] \\
Olmo-3.1-32B & 0.41 [0.35, 0.45] & 0.41 [0.38, 0.45] & 0.39 [0.38, 0.41] & 0.45 [0.43, 0.47] & 0.40 [0.33, 0.47] & 0.42 [0.30, 0.53] & 0.43 [0.26, 0.57] & 0.50 [0.00, 1.00] & 0.41 [0.38, 0.44] \\
Gaperon-24B & 0.50 [0.45, 0.55] & 0.51 [0.47, 0.54] & 0.49 [0.47, 0.50] & 0.46 [0.44, 0.48] & 0.49 [0.43, 0.57] & 0.48 [0.36, 0.61] & 0.60 [0.46, 0.77] & 0.50 [0.00, 1.00] & 0.49 [0.46, 0.51] \\
Gemma-3-27B & 0.74 [0.70, 0.79] & 0.73 [0.70, 0.77] & 0.65 [0.64, 0.67] & 0.73 [0.71, 0.75] & 0.69 [0.63, 0.76] & 0.64 [0.53, 0.74] & 0.77 [0.63, 0.91] & 0.50 [0.00, 1.00] & 0.68 [0.65, 0.71] \\
Granite-30B & 0.59 [0.54, 0.64] & 0.56 [0.53, 0.60] & 0.48 [0.46, 0.49] & 0.56 [0.54, 0.58] & 0.53 [0.45, 0.60] & 0.50 [0.38, 0.62] & 0.66 [0.49, 0.80] & 0.00 [0.00, 0.00] & 0.53 [0.51, 0.56] \\
\hline
\end{tabular}
\end{adjustbox}
\caption{Accuracy and 95\% bootstrap confidence intervals for each model across continents corresponding to Figures \ref{fig:continent_perf} and \ref{fig:continent_perf_full}.}
\label{app:continent_ci}
\end{table*}

%%%%%%%%%%%%%%%%%%%%%%%%%%%%%%%%%%%%%%%%%%%%%%

\section{Additional results by difficulty level, continent, and time period}
\label{app:full_results}

\begin{figure}[h]
    \centering
    \includegraphics[width=\linewidth]{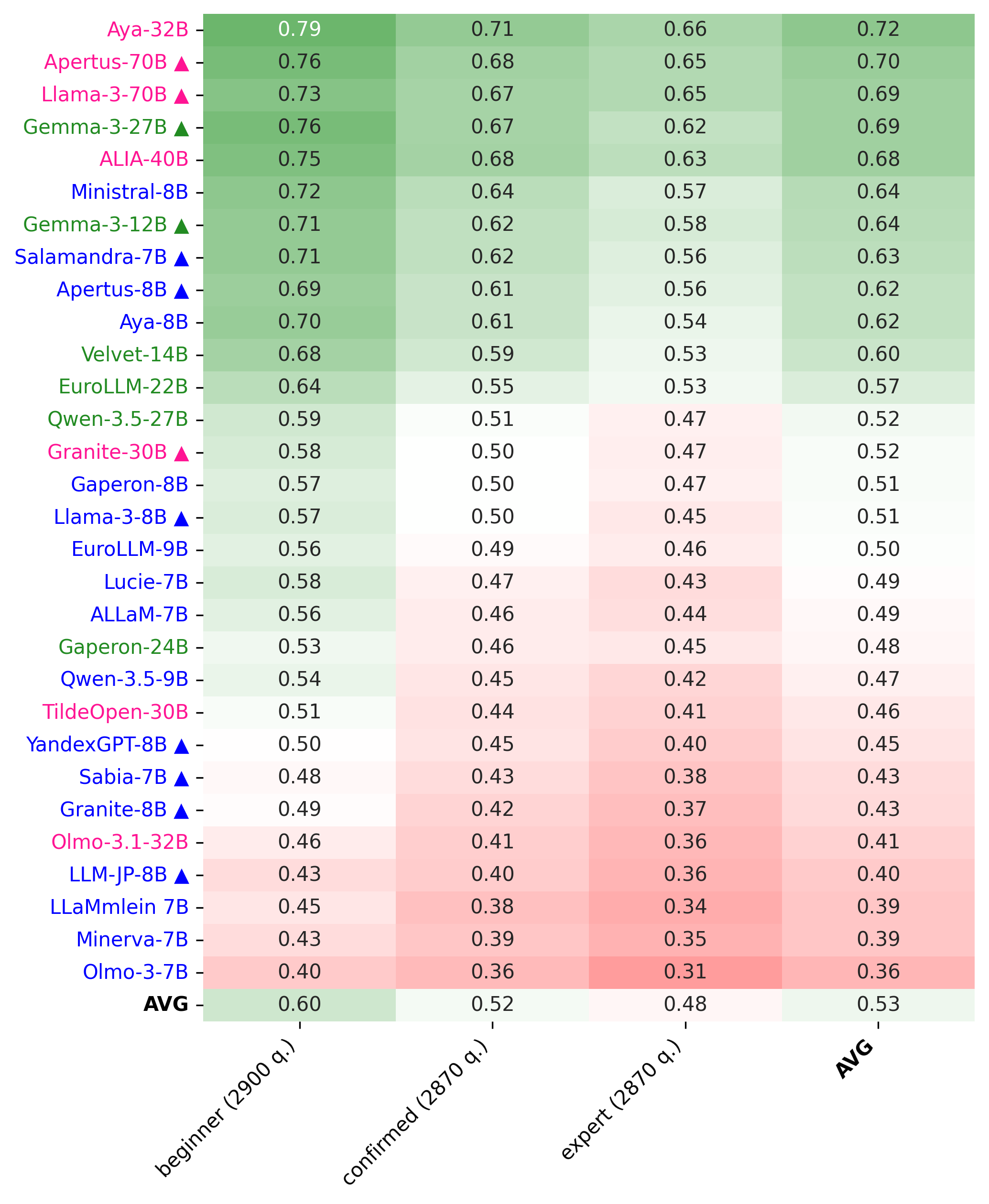}
    \caption{Model performance on \textsc{TriviaRoomQA-French}, grouped by question difficulty. The reported metric is accuracy; greener cells indicate higher scores.}
    \label{fig:detailed_niveau_perf}
\end{figure}

\begin{figure}
    \centering
    \includegraphics[width=\linewidth]{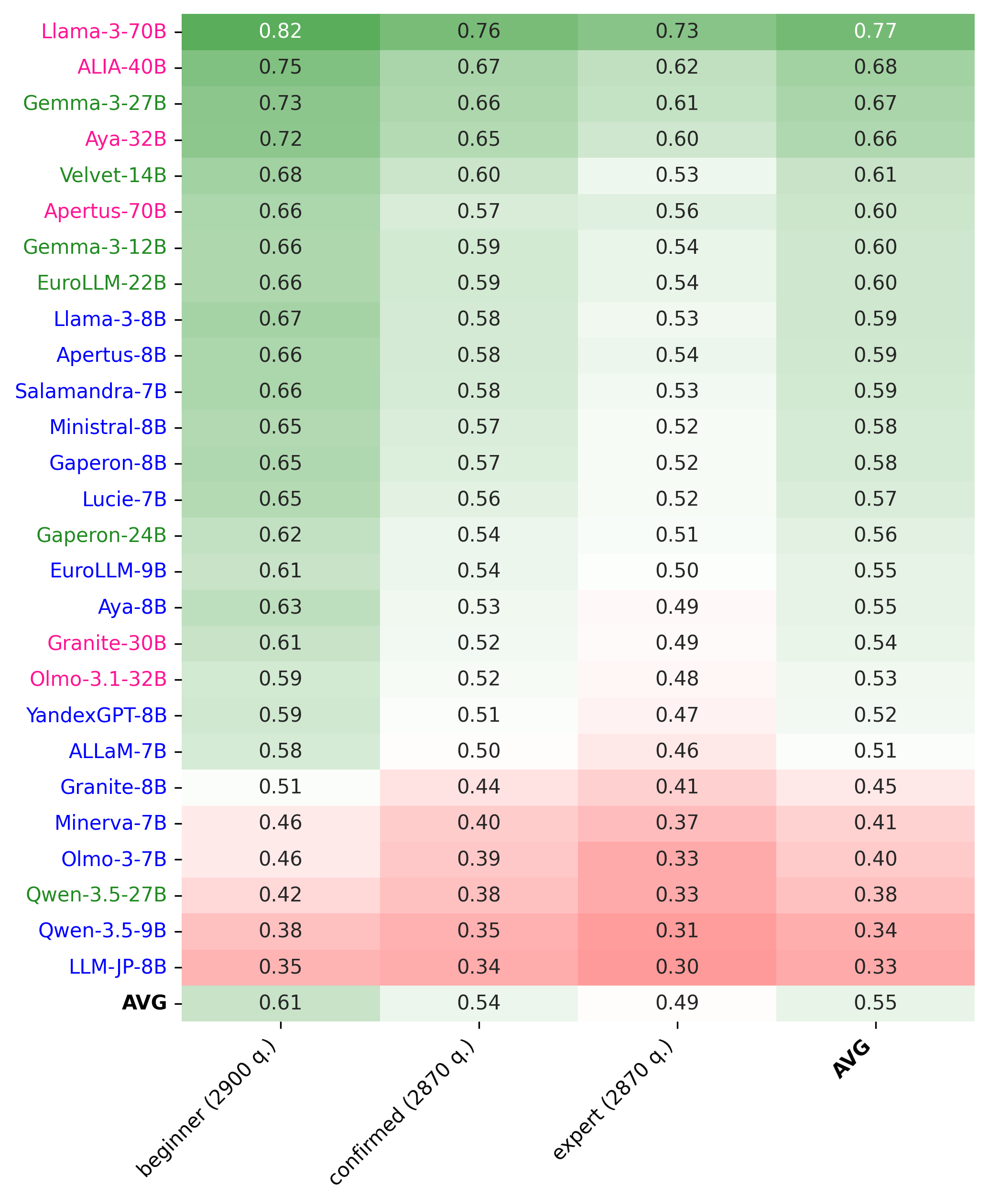}
    \caption{Model performance \textbf{\underline{with chat templates}} on \textsc{TriviaRoomQA-French}, grouped by question difficulty. The reported metric is accuracy.}
    \label{fig:detailed_niveau_perf_chat}
\end{figure}

\begin{figure}[h]
    \centering
    \includegraphics[width=\linewidth]{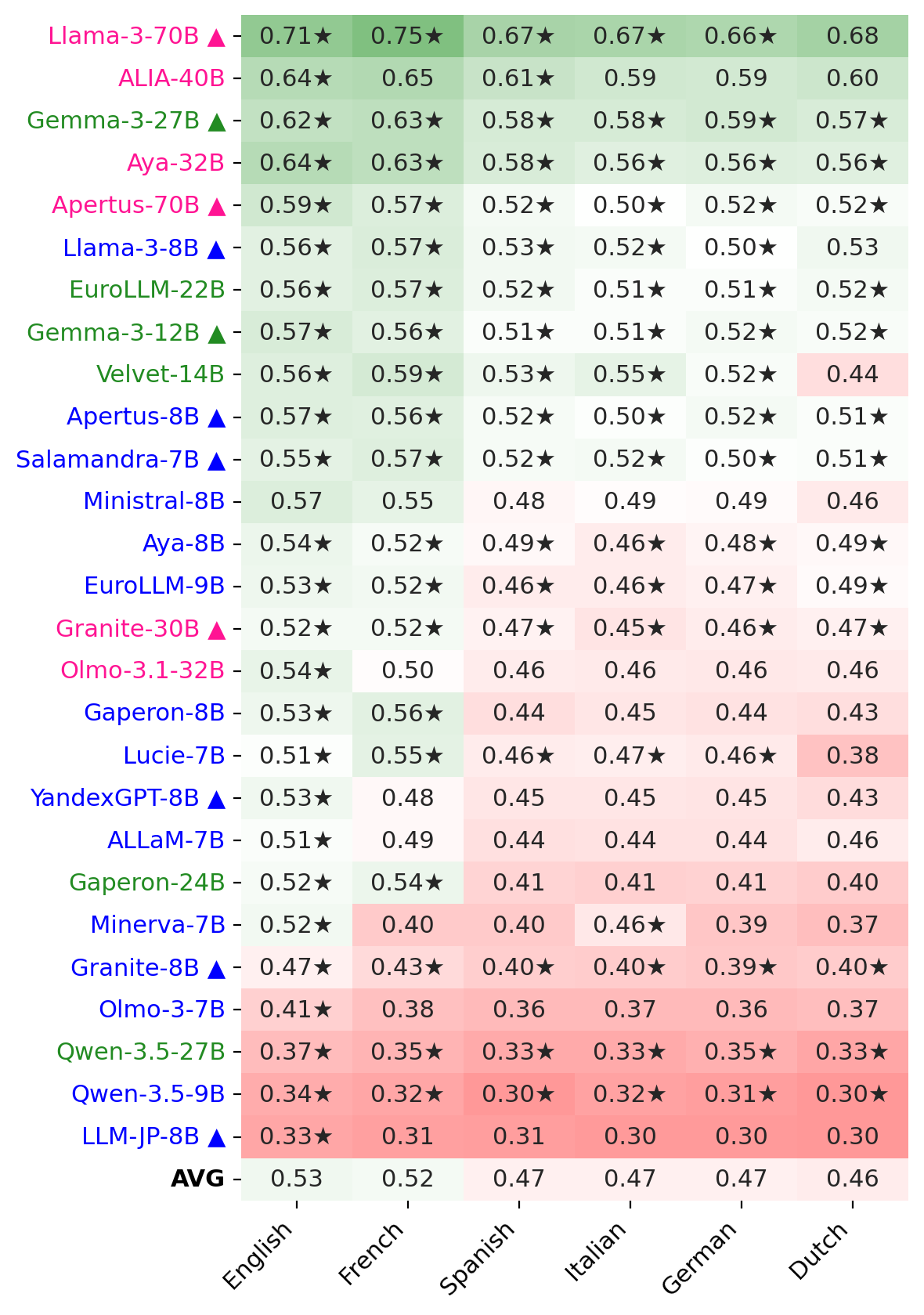}
    \caption{Model performance \textbf{\underline{with chat templates}} on the six languages of \textsc{TriviaRoomQA-Multi}. The reported metric is accuracy.}
    \label{fig:multiling_res_chat}
\end{figure}

\begin{figure*}
    \centering
    \includegraphics[width=1\linewidth]{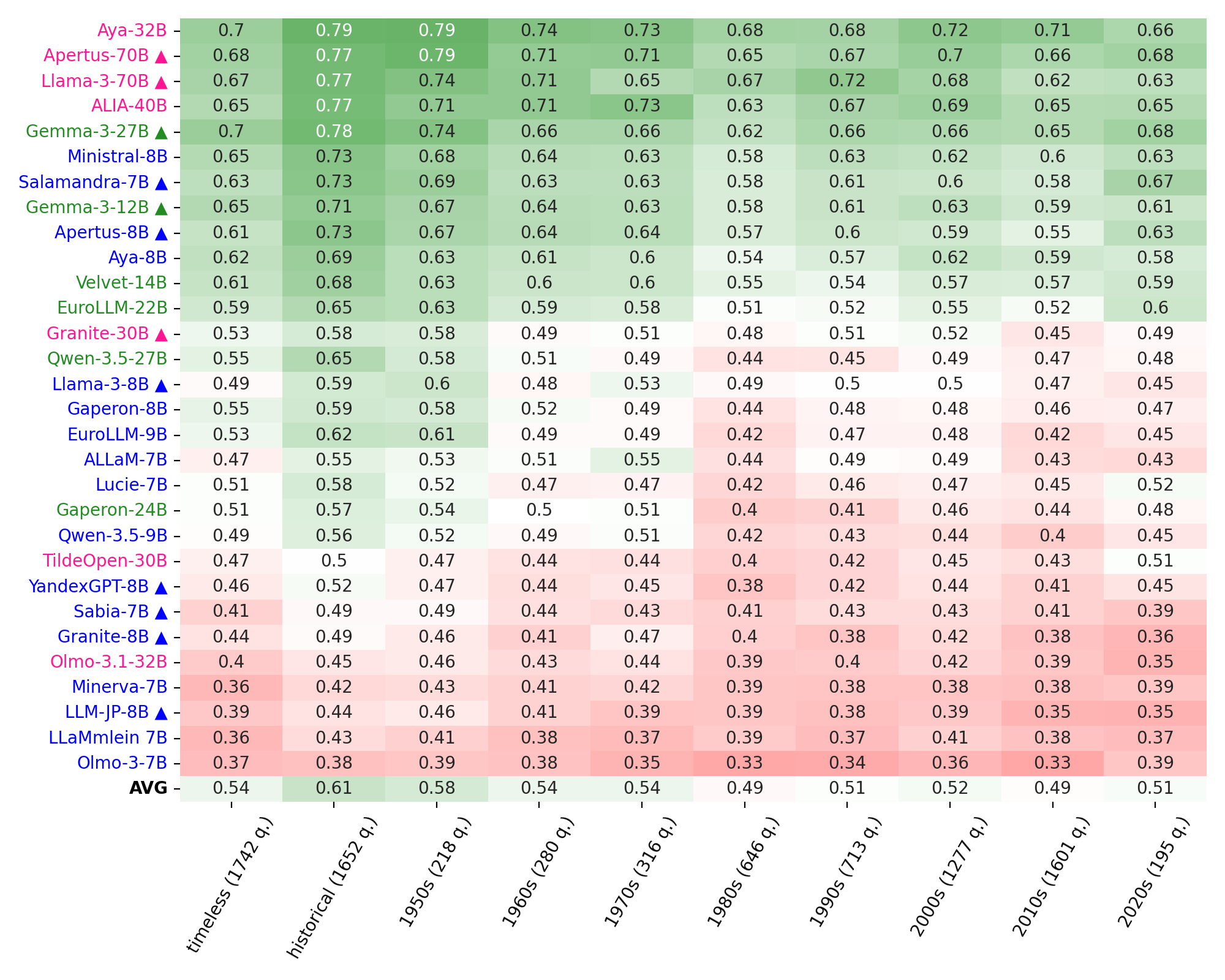}
    \caption{Model performance on \textsc{TriviaRoomQA-French}, grouped by the \textbf{time period} to which each question is most closely related, when applicable. The reported metric is accuracy.}
    \label{fig:timeperiod_perf}
\end{figure*}

\begin{figure*}
    \centering
    \includegraphics[width=1\linewidth]{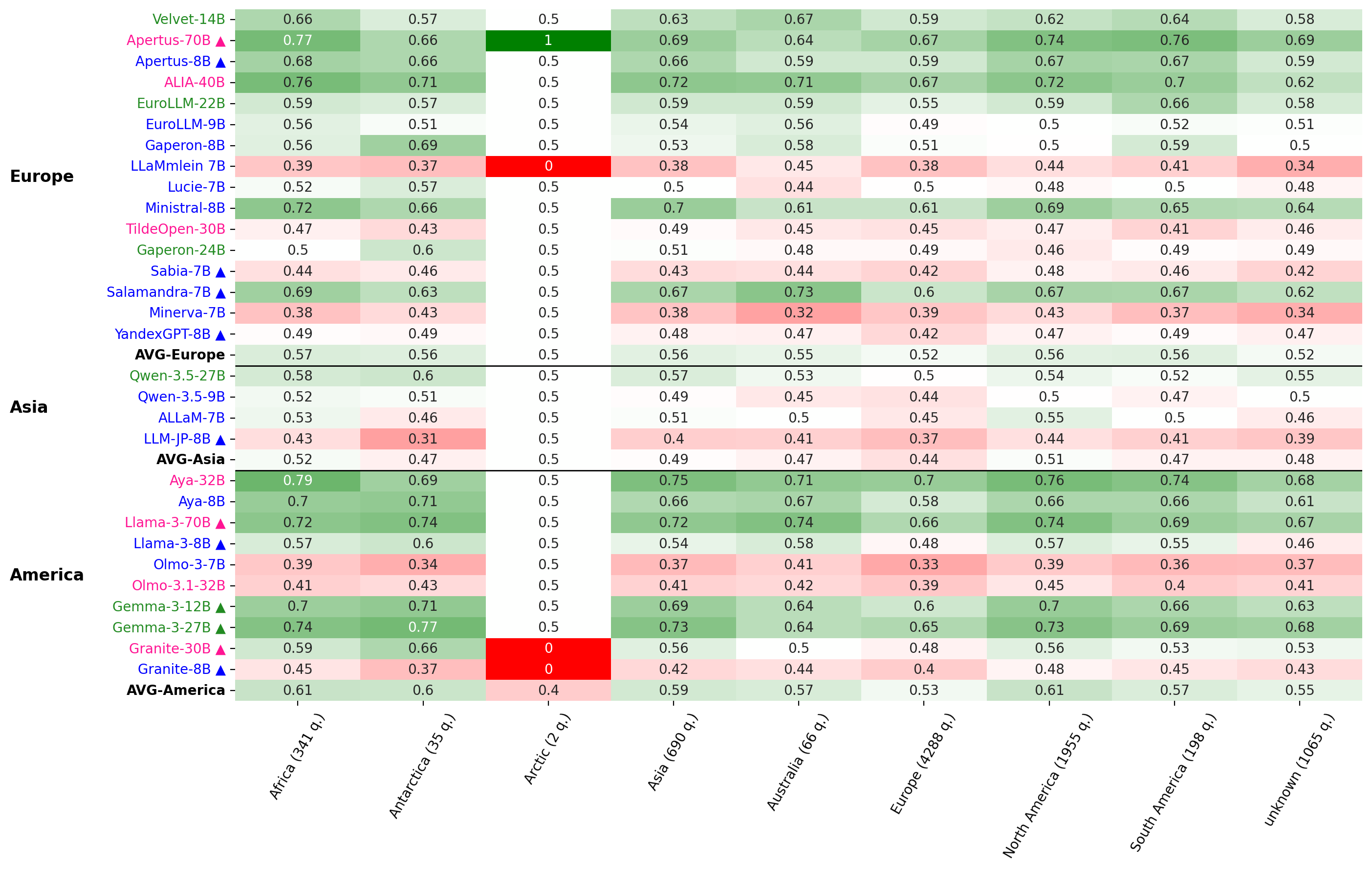}
    \caption{Model performance on \textsc{TriviaRoomQA-French}, grouped by the \textbf{continent} to which each question is most closely related, when applicable. Models are grouped by continent of origin, with the average performance for each group reported in the corresponding \textbf{AVG-\{continent\}} row. The reported metric is accuracy.}
    \label{fig:continent_perf_full}
\end{figure*}

This section provides additional fine-grained results that complement the main analysis. In all figures, the reported metric is accuracy, greener cells indicate higher scores. $\star$ next to a score or a model indicates that the model was trained on that language. $\blacktriangle$ marks models trained on more than 10T tokens. Model size is color-coded: models with less than 10B parameters are shown \textcolor{blue}{in blue}, with 12-27B parameters \textcolor{ForestGreen}{in green}, with 30B+ parameters \textcolor{magenta}{in pink}.
%Models with more than 20B parameters are shown \textcolor{blue}{in blue}, and the remaining models are shown \textcolor{magenta}{in pink}. 

Figure~\ref{fig:detailed_niveau_perf} reports per-model performance by question difficulty, showing that model performance generally follows the annotated difficulty levels: accuracy decreases as questions become harder. Figure~\ref{fig:timeperiod_perf} provides more detailed time-period results, confirming the trend observed in the aggregated results in Figure~\ref{fig:timeperiod_perf_merged}: model performance generally decreases on more contemporary questions. Figure~\ref{fig:continent_perf_full} shows the full continent-level results on \textsc{TriviaRoomQA-French}, including performance on questions for which the continent parameter is not applicable.

\begin{comment}
\begin{figure*}
    \centering
    \includegraphics[width=1\linewidth]{figs/all_lang_per_continent_allmodels.png}
    \caption{Model performance on \textsc{TriviaRoomQA-Multi}, grouped by the \textbf{continent} to which each question is most closely related, when applicable. Results are reported for English, French, Italian, Dutch, German, and Spanish. Models are grouped by continent of origin, with the average performance for each group reported in the corresponding \textbf{AVG-\{continent\}} row. The reported metric is accuracy; greener cells indicate higher scores.}
    \label{fig:all_lang_per_cont}
\end{figure*}
\end{comment}

\section{Results with chat template}
\label{app:chat_template_results}

In the main experiments, we evaluate all models without chat templates to keep the setup comparable across models. In this section, we provide the corresponding results with chat templates for models where this setting is available. In all figures, the reported metric is accuracy, greener cells indicate higher scores. $\star$ next to a score or a model indicates that the model was trained on that language. $\blacktriangle$ marks models trained on more than 10T tokens. Model size is color-coded: models with less than 10B parameters are shown \textcolor{blue}{in blue}, with 12-27B parameters \textcolor{ForestGreen}{in green}, with 30B+ parameters \textcolor{magenta}{in pink}.

\begin{figure*}
    \centering
    \includegraphics[width=\linewidth]{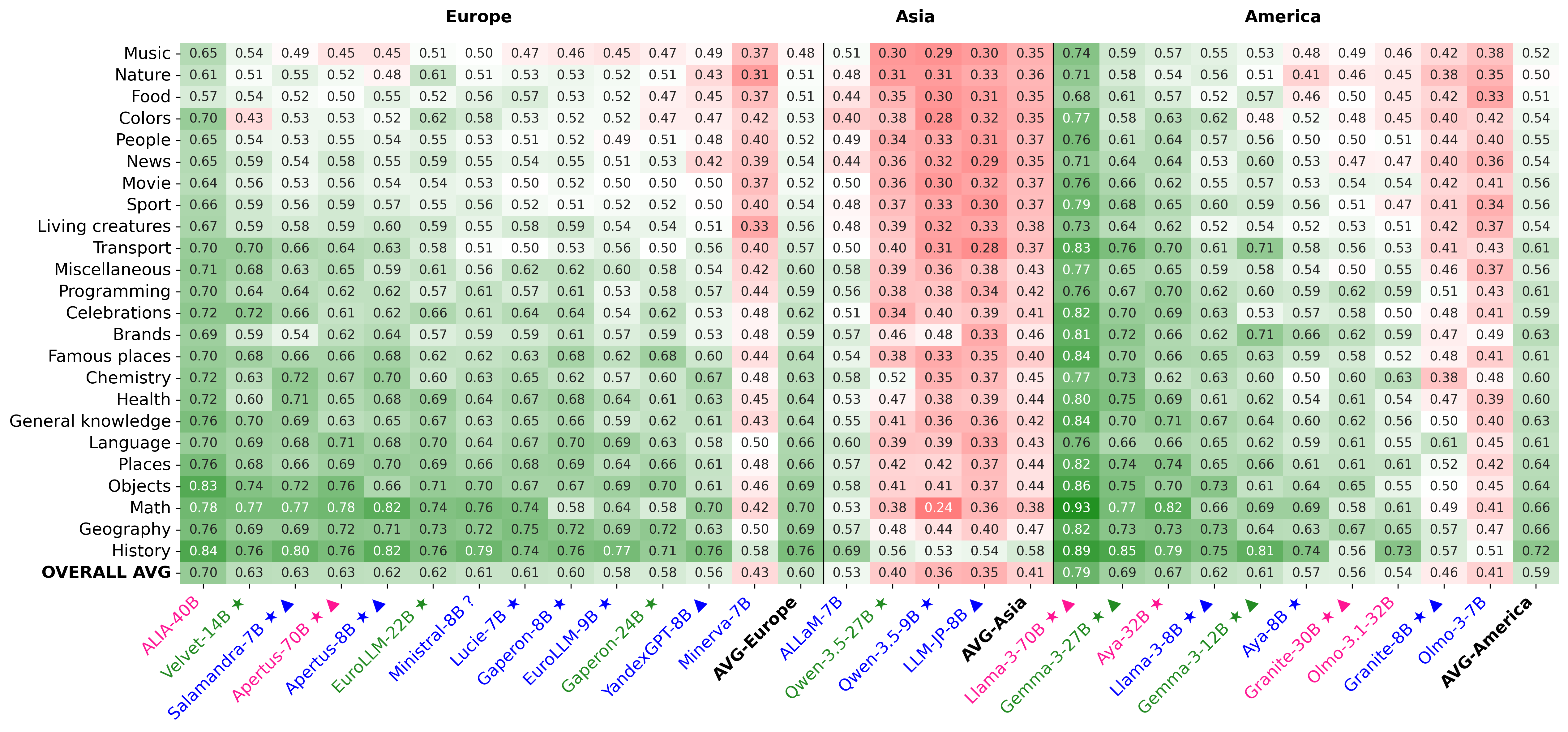}
    \caption{Model performance \textbf{\underline{with chat templates}} on \textbf{24 categories} from \textsc{TriviaRoomQA-French}. The reported metric is accuracy. Models are sorted horizontally from left to right by performance within each region, and categories are sorted vertically from top to bottom by average performance across all models.}
    \label{fig:detailed_category_chat}
\end{figure*}

\begin{figure*}
    \centering
    \includegraphics[width=1\linewidth]{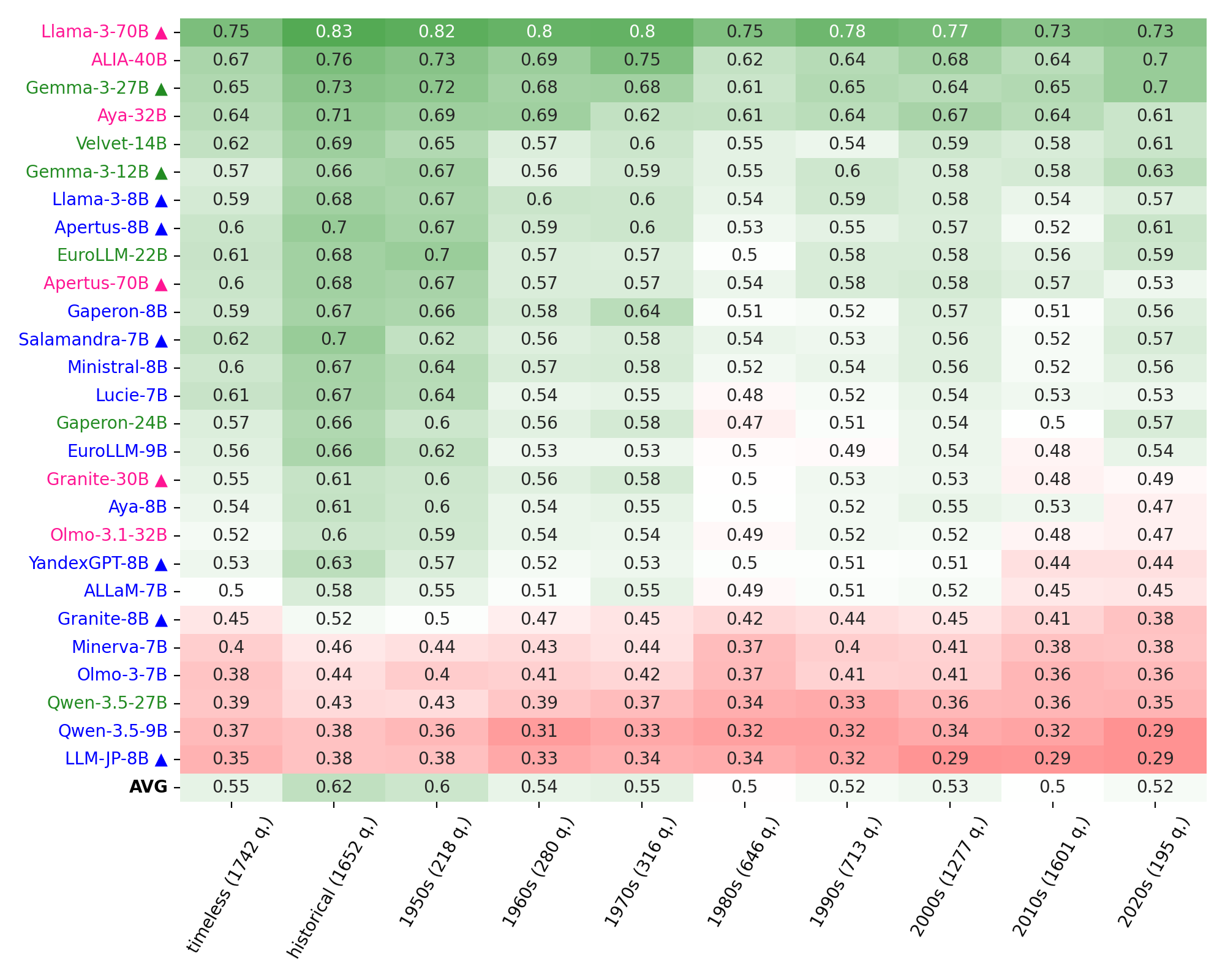}
    \caption{Model performance \textbf{\underline{with chat templates}} on \textsc{TriviaRoomQA-French}, grouped by the \textbf{time period} to which each question is most closely related, when applicable. The reported metric is accuracy.}
    \label{fig:timeperiod_perf_chat}
\end{figure*}

\begin{figure*}
    \centering
    \includegraphics[width=1\linewidth]{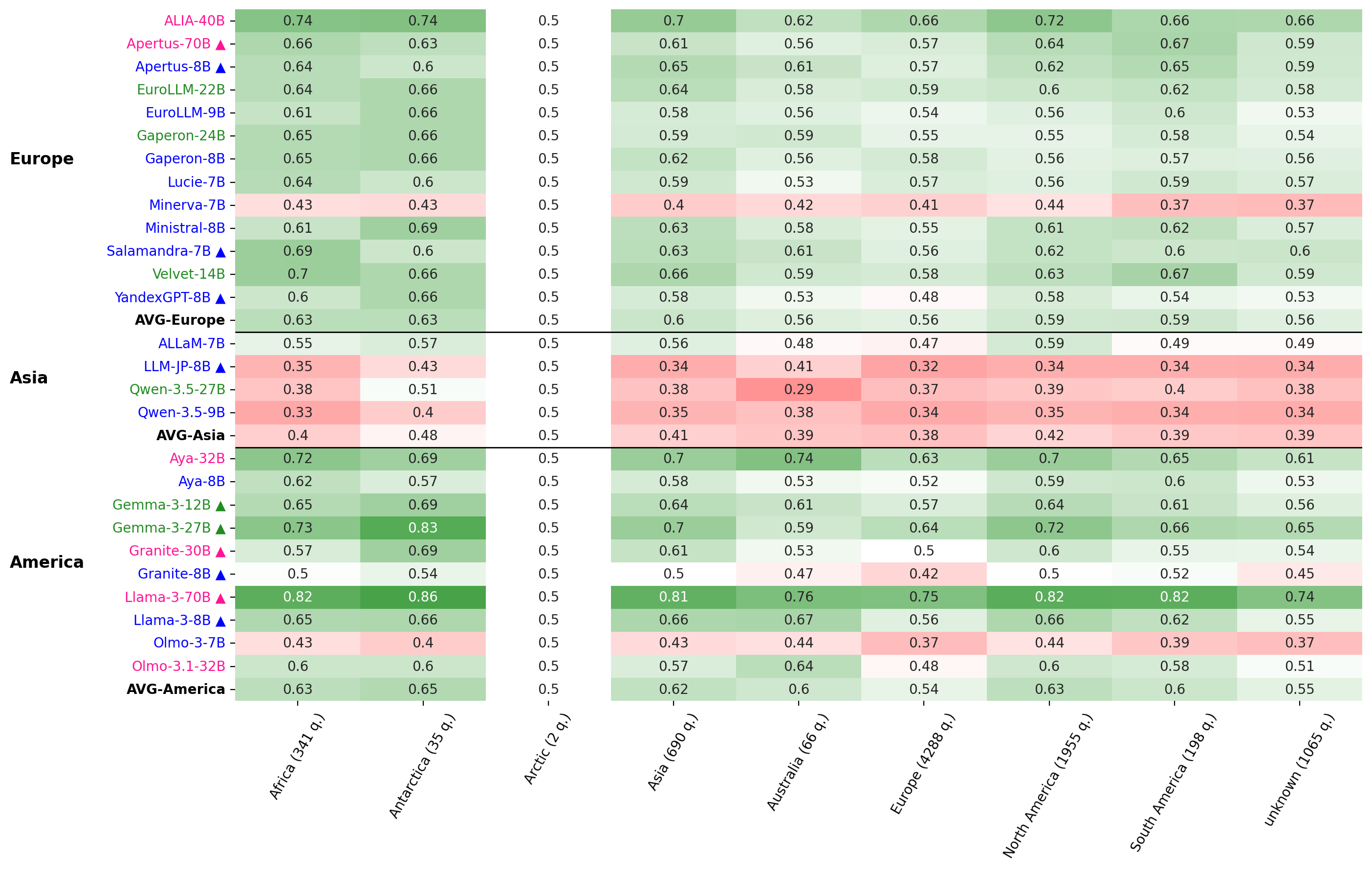}
    \caption{Model performance \textbf{\underline{with chat templates}} on \textsc{TriviaRoomQA-French}, grouped by the \textbf{continent} to which each question is most closely related, when applicable. Models are grouped by continent of origin, with the average performance for each group reported in the corresponding \textbf{AVG-\{continent\}} row. The reported metric is accuracy.}
    \label{fig:continent_perf_chat}
\end{figure*}

Figure~\ref{fig:multiling_res_chat} reports performance on the multilingual setting, while Figures~\ref{fig:detailed_niveau_perf_chat}, \ref{fig:detailed_category_chat}, \ref{fig:timeperiod_perf_chat}, and~\ref{fig:continent_perf_chat} report results on the French collection by category, difficulty level, time period, and continent, respectively. Overall, the main trends remain similar to the no-chat-template setting: performance varies across languages, encyclopedic categories remain easier than popular-culture categories, and models still show variation across time periods and geographic regions.

\begin{table*}[t]
\centering
\small
\begin{tabularx}{\textwidth}{p{0.15\textwidth} X}
\toprule
\textbf{Category} & \textbf{Topics} \\
\midrule
Brands & Brands, logos and slogans; Coca-Cola Company; Microsoft \\
Celebrations & Merry Christmas; Halloween; Saint Nicholas Day \\
Chemistry & Magnesium; Chemistry \\
Places & Canada; Italy; Nice; Belgium; United Kingdom; Rome; Brittany; Cities of the world; Vannes; Auroville; Périgord; Mouscron; City nicknames; Brussels today; Istanbul; Japan \\
Colors & Colors; Full of color \\
General knowledge & International knowledge; Knowledge and personalities; Folklore; Romanticism; Saints; General knowledge 2; General knowledge 3; General knowledge 4; General knowledge 5; Mixed knowledge; Mixed knowledge 2; Unbeatable; Youth knowledge; Mixed knowledge 3; Mixed knowledge 4; Mixed knowledge 5; Mixed knowledge 6; Japanese folklore; Sculpture; General knowledge \\
Famous places & The call of the open sea; Machu Picchu; Teotihuacan; World monuments; Castles and fortified castles; Central Park; Mont Saint-Michel; Tourist sites; Very long bridges; Louvre Museum; Chambord \\
Food & Desserts and pastries; Wines from elsewhere; Sugar; Coffee; Chocolate; Breakfast; Gin; Potato; Herbs and spices; Mineral waters; Belgian beers; Apples; Foreign gastronomy; French cheeses; Divine wines; Non-alcoholic drinks \\
Geography & Mediterranean; Geography for everyone; The Durance; Water stories; Antarctica; Active volcanoes; Forests of France; Long-distance hikes \\
Health & The human body; Health and well-being; The Omicron threat; Hair \\
History & Ancient Egypt; Political history; History of France; Gladiators; Wars and battles; Major dates of the 20th century \\
Language & Literary quotations; Famous expressions; Short quotations; Internet Frenglish; Spelling quiz; Crosswords; Crosswords 2 \\
Living creatures & Sharks; Famous animals; Birds; Horses; Animals in numbers; Animals of all kinds; Beehive bees; Pigeon racing; Ants; Animals and habitats; Cats \\
Movie & Pokémon; SpongeBob SquarePants; Tintin; Movie robots; Television serials; Playboy; French comedies; Youth and cartoons; Star Trek; Secret Story; Disney princesses; Superheroines; Reality TV; César Awards ceremony; Alien: the saga; Fiction for everyone; Comedy films; Money Heist; Harry Potter; The X-Files; Breaking Bad; Marvel heroes; Game of Thrones; The Fifth Element; American series; The Big Blue; The Lion King; The Visitors; Jurassic Park; The Hunger Games; Toy Story turns 20; Star Wars heroes; Star Wars; Movie secrets; Imaginary characters \\
Music & Eighties bands; French variety music; The Cure; Disco hits; New wave; Reggae; Acid jazz \\
Nature & Fruit trees; Snow; Cactus; Japanese garden; Field fauna and flora \\
News & Celebrity news: February 2017; Celebrity news: January 2017; Celebrity news: March 2017; Celebrity news: April 2017; Celebrity news: May 2017; Celebrity news: June 2017; Celebrity news: August 2017; Celebrity news: September 2017; Social issues; Celebrity news: October 2017; Celebrity news: January 2018; Celebrity news: February 2018; Celebrity news: March 2018; Celebrity news: April 2018; Celebrity news: May 2018; Celebrity news: June 2018; Celebrity news: July 2018; Celebrity news: August 2018; Celebrity news: September 2018; Celebrity news: October 2018; It happened in 2019; COVID-19; Celebrity news: February 2021; Celebrity news: March 2021; Celebrity news: April 2021; Celebrity news: May 2021; Celebrity news: June 2021; 2021 retrospective; Celebrity news: August 2015; Celebrity news: February 2015; Celebrity news: January 2015; Celebrity news: September 2015 \\
Math & Find the number; Questions about numbers; Units of measurement \\
Objects & Inventions; Objects and instruments; Makeup; Ceramics and pottery; Musical instruments \\
Miscellaneous & Global warming; Fashion victim; Mysteries of the world; Dragons yesterday and today; Board games; Languages of love; Peoples of the world \\
People & Mothers; Depeche Mode; Jenna Jameson; Clara Morgane; Donald Trump; Classical authors; Maria Sharapova; Mythical movie couples; Nikola Tesla; Steven Spielberg; Handsome men; Johnny Hallyday; Mike Horn; France Gall; Alan Turing; Global stars; Meghan Markle; Famous first names; WAGs; Traci Lords; Britney Spears; Brad Pitt in film; Pamela Anderson; French TV hosts; Florence Foresti; Charlize Theron; John McEnroe; Famous Alberts; Electro artists; Celebrity gossip 2013; Jean-Marie Bigard; Virginie on screen; Patrick Sébastien; Clint Eastwood; Celebrity gossip 2014; Cinema beauties; Bruce Willis; Sophie Marceau; French female directors; Jean Michel Jarre; Marilyn Monroe; Celebrities; Tutankhamun; Maxime Chattam; International singers \\
Programming & Geocaching; Internet; PlayStation 2; Cryptocurrencies; OpenBSD; Instagram; World of Warcraft; Software and web applications; Nintendo games and consoles; Linux; iPhone; Social networks \\
Sport & Rio 2016; Weightlifting; Boxing; Football of the past; Sports for everyone; Ironman; Ligue 1 stadiums; Climbing; PyeongChang 2018; Golf; Tennis; Russia 2018; Olympic Games; Football 2010--2020; Football 2000--2010; Football 1990--2000; Tennis court; Team sports; NBA players and franchises; European basketball; FC Barcelona \\
Transport & Means of transport; Car manufacturers; Automobile \\
\bottomrule
\end{tabularx}
\caption{Categories and associated topics in \textsc{TriviaRoomQA-French}.}
\label{tab:categories_topics}
\end{table*}

\begin{table*}[t]
\centering
\small
\begin{tabular}{p{0.96\linewidth}}
\hline
\textbf{Continent classification prompt} \\
\hline
You are a continent classifier. \\[0.5em]

Task: \\
Identify the continent most strongly associated with the quiz question. \\[0.5em]

You may use: \\
- the question \\
- the candidate answers \\
- the anecdote \\
- the quiz topic/category \\[0.5em]

Allowed outputs ONLY: \\
- Africa \\
- Antarctica \\
- Arctic \\
- Asia \\
- Europe \\
- North America \\
- Australia \\
- South America \\
- unknown \\[0.5em]

Rules: \\
- Output ONLY one valid continent name from the allowed list. \\
- If the question is not related to a specific continent, output: unknown \\
- Never answer the quiz question itself. \\
- Never explain. \\
- Never ask for clarification. \\
- Never output countries, cities, companies, objects, animals, or sentences. \\
- Output exactly one value from the allowed list. \\[0.5em]

Examples: \\[0.5em]

Question: Quelle marque de jeans mondialement connue fabrique le célèbre 501 ? \\
Choices: Levi's, Wrangler, Diesel, Lee \\
Anecdote: Cette compagnie de vêtements est en effet mondialement connue, principalement pour son Blue-jeans. \\
Topic: Marques et entreprises \\
Continent: North America \\[0.5em]

Question: De quelle filiale du groupe Nestlé George Clooney est-il devenu le séduisant ambassadeur ? \\
Choices: Nespresso, Perrier, Maggi, KitKat \\
Anecdote: Le fameux slogan `What else ?', associé au beau George Clooney, a aujourd'hui été souvent parodié et a fait le tour du monde. \\
Topic: Publicité \\
Continent: Europe \\[0.5em]

Question: Grâce à quoi les requins peuvent-ils respirer, de la même manière que les raies ? \\
Choices: Leurs poumons, Leurs nageoires, Leurs branchies, Leur peau \\
Anecdote: Les requins possèdent cinq à sept fentes branchiales sur les côtés de la tête. \\
Topic: Animaux \\
Continent: unknown \\[0.5em]

Question: \{context\} \\
Choices: \{", ".join(answers)\} \\
Anecdote: \{anecdote\} \\
Topic: \{topic\} \\
Continent: \\
\hline
\end{tabular}
\caption{Prompt used for continent classification.}
\label{tab:continent-prompt}
\end{table*}

\begin{table*}[t]
\centering
\small
\begin{adjustbox}{max width = 0.9\textwidth}
\begin{tabular}{p{0.96\linewidth}}
\hline
\textbf{Time-period classification prompt} \\
\hline
You are a quiz question time-period classifier. \\[0.5em]

Task: \\
Identify the decade or era most strongly associated with the subject of the quiz question. \\[0.5em]

You may use: \\
- the question \\
- the candidate answers \\
- the anecdote \\
- the quiz topic/category \\[0.5em]

Allowed outputs ONLY: \\
- 2020s \\
- 2010s \\
- 2000s \\
- 1990s \\
- 1980s \\
- 1970s \\
- 1960s \\
- 1950s \\
- historical \\
- timeless \\[0.5em]

Rules: \\
- Output ONLY one valid value from the allowed list. \\
- Select the period most directly associated with the main subject of the question. \\
- Use the period during which the referenced person, event, product, company, work, trend, or phenomenon is most culturally relevant or famous. \\
- For questions about discoveries, inventions, brands, celebrities, songs, movies, politics, sports, or technology, choose the decade most associated with them. \\
- For questions about ancient history, the Middle Ages, Antiquity, monarchies, wars before 1950, mythology, or old civilizations, output: historical \\
- For questions about biology, mathematics, language, geography, nature, physics, animals, or other knowledge not tied to a specific era, output: timeless \\
- If multiple periods are possible, choose the one that is most commonly associated with the topic. \\
- Never answer the quiz question itself. \\
- Never explain. \\
- Never ask for clarification. \\
- Never output anything other than one allowed value. \\
- Output exactly one value. \\[0.5em]

Examples: \\[0.5em]

Question: Quelle marque de jeans mondialement connue fabrique le célèbre 501 ? \\
Choices: Levi's, Wrangler, Diesel, Lee \\
Anecdote: Cette compagnie de vêtements est en effet mondialement connue, principalement pour son Blue-jeans. \\
Topic: Marques et entreprises \\
Time period: historical \\[0.5em]

Question: De quelle filiale du groupe Nestlé George Clooney est-il devenu le séduisant ambassadeur ? \\
Choices: Nespresso, Perrier, Maggi, KitKat \\
Anecdote: Le fameux slogan `What else ?', associé au beau George Clooney, a aujourd'hui été souvent parodié et a fait le tour du monde. \\
Topic: Publicité \\
Time period: 2000s \\[0.5em]

Question: Grâce à quoi les requins peuvent-ils respirer, de la même manière que les raies ? \\
Choices: Leurs poumons, Leurs nageoires, Leurs branchies, Leur peau \\
Anecdote: Les requins possèdent cinq à sept fentes branchiales sur les côtés de la tête. \\
Topic: Animaux \\
Time period: timeless \\[0.5em]

Question: Quelle ville française était la « cité des papes » au Moyen-Âge ? \\
Choices: Metz, Nantes, Tours, Avignon \\
Anecdote: C'est l'une des rares villes françaises à avoir conservé ses remparts et son centre historique, composé du palais des papes. \\
Topic: Histoire de France (Paris vaut bien une messe) \\
Time period: historical \\[0.5em]

Question: \{context\} \\
Choices: \{", ".join(answers)\} \\
Anecdote: \{anecdote\} \\
Topic: \{topic\} \\
Time period: \\
\hline
\end{tabular}
\end{adjustbox}
\caption{Prompt used for time-period classification.}
\label{tab:time-period-prompt}
\end{table*}

\section{Model details}

This section provides additional information about the models used in our experiments. Table~\ref{tab:models_training_data} summarizes model-level metadata, including the provider region, reported training data scale, and language coverage. Table~\ref{tab:model_references_urls} provides the corresponding model references and URLs. When available, we rely on information reported in the model papers or official model cards, missing values indicate that the corresponding information was not explicitly reported by the model providers.

\section{GPU usage}

For most evaluations described in this paper, we used a single Nvidia H100 80GB GPU. The exceptions were the evaluations of Apertus-70B-Instruct-2509 and Llama-3.1-70B-Instruct, for which we used two H100 cards. In total, evaluating the two 70B models took 30 minutes without the chat template, while evaluating the remaining 28 models took 3 hours. The same amount of time was required for the evaluations with the chat template. 

The evaluation of search-augmented models, Llama-3.1-8B-Instruct and Qwen3.5-9B, also used one H100 GPU. It took 4 hours for Llama-3.1-8B-Instruct and 30 hours for Qwen3.5-9B.

\section{Experimental setup details}
\label{app:exp_details}

For evaluation, we used the LM Evaluation Harness \citep{eval-harness}\footnote{https://github.com/EleutherAI/lm-evaluation-harness}, for which we implemented two new tasks as YAML configuration files. All models were accessed through the Transformers library \citep{wolf-etal-2020-transformers}. We evaluated each model once with a batch size of 16 and zero few-shot examples, under two conditions: with and without the chat template. Unless otherwise specified, all other parameters were left at their default values. We report average model's accuracy aggregated along different axes, including category, language, time period, continent, and difficulty level.

Experiments were run using LM Evaluation Harness v0.4.11 and Transformers v5.5.4.

\begin{table*}[]
\centering
\begin{adjustbox}{max width = 0.95\textwidth}
\begin{tabular}{p{4.6cm}p{2cm}cp{5.5cm}}
\hline
\textbf{Model name} & \textbf{\# Training Tokens} & \textbf{Country} & \textbf{Languages} \\
\hline

\multicolumn{4}{c}{\textbf{Europe}} \\

\hline

EuroLLM-9B-Instruct & 4T & European Union & {35 languages (Dutch, English, French, German, Italian, Spanish)} \\
EuroLLM-22B-Instruct-2512 & 4T &  & \\
\hline
sabia-7b & 11.4T & Portugal & English, Portuguese \\
\hline
ALIA-40b-instruct-2601 & 9.37T & Spain & Dutch, English, French, German, Italian, Spanish \\
\hline
salamandra-7b-instruct & 12.875T & Spain & 35 languages (Dutch, English, French, German, Italian, Spanish) \\
\hline
Lucie-7B-Instruct-v1.1 & 3T & France & English, French, German, Italian, Spanish \\
\hline
Ministral-8B-Instruct-2410 & - & France & - \\
\hline
Gaperon-1125-8B-SFT & 4T & France & \multirow{2}{*}{English, French} \\
Gaperon-1125-24B-SFT & 2T &  & \\
\hline
sapienzanlp--Minerva-7B-instruct-v1.0 & 2.5T & Italy & English, Italian \\
\hline
Velvet-14B & 4T & Italy & English, French, German, Italian, Portuguese, Spanish \\
\hline
Apertus-8B-Instruct-2509 & 15T & Switzerland & {1800 languages (English, French, German, Italian, Portuguese, Spanish)} \\
Apertus-70B-Instruct-2509 & 15T &  & \\
\hline
LLaMmlein-7B & 3T$^\star$ & Germany & German \\
\hline
TildeAI--TildeOpen-30b & 2T & Lithuania & 34 languages (Dutch, English, French, German, Italian, Spanish) \\
\hline
YandexGPT-5-Lite-8B-instruct & 15.32T & Russia & English, Russian \\
\hline
\multicolumn{4}{c}{\textbf{America}} \\
\hline
Llama-3.1-8B-Instruct & 15T & USA & 8 languages (English, French, German, Italian, Spanish) \\
Llama-3.1-70B-Instruct & 15T & & \\
\hline
granite-4.1-8b & 15T & USA & {Multilingual (includes Dutch, English, French, German, Italian, Spanish)} \\
granite-4.1-30b & 15T &  & \\
\hline
gemma-3-12b-it & 12T & USA & \multirow{2}{*}{140 languages} \\
gemma-3-27b-it & 14T &  & \\
\hline
Olmo-3-7B-Instruct & 6.05T & USA & \multirow{2}{*}{English} \\
Olmo-3.1-32B-Instruct & 6.2T &  & \\
\hline
aya-expanse-8b & - & Canada & 23 languages (Dutch, English, French, German, Italian, Spanish) \\
aya-expanse-32b & - &  & \\
\hline

\multicolumn{4}{c}{\textbf{Asia}} \\
\hline
ALLaM-7B-Instruct-preview & 5.2T & Saudi Arabia & Arabic, English \\
\hline
llm-jp--llm-jp-4-8b-instruct & 11.7T & Japan & English, Japanese \\
\hline
Qwen3.5-9B & - & China & 200+ languages \\
Qwen3.5-27B & - &  &  \\
\hline

\end{tabular}
\end{adjustbox}
\caption{Training data details for the models used in the evaluation, including provider country and language coverage. $^\star$The authors of  LLaMmlein \citep{pfister-etal-2025-llammlein} do not explicitly report the training dataset size for LLaMmlein-7B, but state that the same training dataset was used for all model sizes (LLaMmlein 120M, 1B, 7B). We therefore report the number of training tokens given for the LLaMmlein-1B model.}
\label{tab:models_training_data}
\end{table*}

\begin{table*}[]
\centering
\begin{tabular}{p{5cm}p{10cm}}
\hline
\textbf{Model name} & \textbf{URL} \\
\hline
EuroLLM-9B-Instruct \citep{martins2025eurollm9btechnicalreport} & \small{\url{https://huggingface.co/utter-project/EuroLLM-9B-Instruct}} \\
EuroLLM-22B-Instruct-2512 \citep{ramos2026eurollm22btechnicalreport} & \small{\url{https://huggingface.co/utter-project/EuroLLM-22B-Instruct-2512}} \\
\hline
sabia-7b \citep{Pires_2023} & \small{\url{https://huggingface.co/maritaca-ai/sabia-7b}} \\
\hline
ALIA-40b-instruct-2601 \citep{gonzalezagirre2025salamandratechnicalreport} & \small{\url{https://huggingface.co/BSC-LT/ALIA-40b-instruct-2601}} \\
\hline
salamandra-7b-instruct \citep{gonzalezagirre2025salamandratechnicalreport} & \small{\url{https://huggingface.co/BSC-LT/salamandra-7b-instruct}} \\
\hline
Lucie-7B-Instruct-v1.1 \citep{gouvert2025lucie7bllmlucietraining} & \small{\url{https://huggingface.co/OpenLLM-France/Lucie-7B-Instruct-v1.1}} \\
\hline
Ministral-8B-Instruct-2410 & \small{\url{https://huggingface.co/mistralai/Ministral-8B-Instruct-2410}} \\
\hline
Gaperon-1125-8B-SFT \citep{godey2025gaperonpepperedEnglishfrenchgenerative} & \small{\url{https://huggingface.co/almanach/Gaperon-1125-8B-SFT}} \\
Gaperon-1125-24B-SFT  & \small{\url{https://huggingface.co/almanach/Gaperon-1125-24B-SFT}} \\
\hline
sapienzanlp--Minerva-7B-instruct-v1.0 \citep{orlando-etal-2024-minerva} & \small{\url{https://huggingface.co/sapienzanlp/Minerva-7B-instruct-v1.0}} \\
\hline
Velvet-14B & \small{\url{https://huggingface.co/Almawave/Velvet-14B}} \\
\hline
Apertus-8B-Instruct-2509 \citep{apertus2025apertusdemocratizingopencompliant} & \small{\url{https://huggingface.co/swiss-ai/Apertus-8B-Instruct-2509}} \\
Apertus-70B-Instruct-2509  & \small{\url{https://huggingface.co/swiss-ai/Apertus-70B-Instruct-2509}} \\
\hline
LLaMmlein-7B \citep{pfister-etal-2025-llammlein} & \small{\url{https://huggingface.co/LSX-UniWue/LLaMmlein_7B}} \\
\hline
TildeAI--TildeOpen-30b \citep{bergmanis2026tildeopenllmleveragingcurriculum} & \small{\url{https://huggingface.co/TildeAI/TildeOpen-30b}} \\
\hline
YandexGPT-5-Lite-8B-instruct & \small{\url{https://huggingface.co/yandex/YandexGPT-5-Lite-8B-instruct}} \\
\hline
Llama-3.1-8B-Instruct \citep{grattafiori2024llama3herdmodels} & \small{\url{https://huggingface.co/meta-llama/Llama-3.1-8B-Instruct}} \\
Llama-3.1-70B-Instruct  & \small{\url{https://huggingface.co/meta-llama/Llama-3.1-70B-Instruct}} \\
\hline
granite-4.1-8b & \small{\url{https://huggingface.co/ibm-granite/granite-4.1-8b}} \\
granite-4.1-30b & \small{\url{https://huggingface.co/ibm-granite/granite-4.1-30b}} \\
\hline
gemma-3-12b-it \citep{gemma_2025} & \small{\url{https://huggingface.co/google/gemma-3-12b-it}} \\
gemma-3-27b-it  & \small{\url{https://huggingface.co/google/gemma-3-27b-it}} \\
\hline
Olmo-3-7B-Instruct \citep{olmo2026olmo3} & \small{\url{https://huggingface.co/allenai/Olmo-3-7B-Instruct}} \\
Olmo-3.1-32B-Instruct  & \small{\url{https://huggingface.co/allenai/Olmo-3.1-32B-Instruct}} \\
\hline
aya-expanse-8b \citep{dang2024ayaexpansecombiningresearch} & \small{\url{https://huggingface.co/CohereLabs/aya-expanse-8b}} \\
aya-expanse-32b  & \small{\url{https://huggingface.co/CohereLabs/aya-expanse-32b}} \\
\hline
ALLaM-7B-Instruct-preview \citep{bari2025allam} & \small{\url{https://huggingface.co/humain-ai/ALLaM-7B-Instruct-preview}} \\
\hline
llm-jp--llm-jp-4-8b-instruct & \small{\url{https://huggingface.co/llm-jp/llm-jp-4-8b-instruct}} \\
\hline
Qwen3.5-9B \citep{qwen3.5} & \small{\url{https://huggingface.co/Qwen/Qwen3.5-9B}} \\
Qwen3.5-27B  & \small{\url{https://huggingface.co/Qwen/Qwen3.5-27B}} \\
\hline
\end{tabular}
\caption{Model references and URLs. When multiple model sizes come from the same paper, the citation is included only once.}
\label{tab:model_references_urls}
\end{table*}

\end{document}